%% file: paper.tex
\documentclass{article}

\usepackage{microtype}
\usepackage{graphicx}
\usepackage{subcaption}
\usepackage{booktabs} 
\usepackage{siunitx} 
\usepackage{enumitem}
\usepackage{tikz}
\usepackage{placeins}

\usepackage{hyperref}

\usepackage[accepted]{icml2019}

\input{math_commands}

\newcommand{\cCT}{c_\text{CT}}
\newcommand{\cSSL}{c_\text{S\textsuperscript{2}L}}
\newcommand{\cCL}{c_\text{CL}}
\newcommand{\ySSL}{\hat y_\text{S\textsuperscript{2}L}}
\newcommand{\yCL}{\hat y_\text{CL}}
\newcommand{\cRF}{c_\text{r/f}}
\newcommand{\cRot}{c_\text{R}}

\newcommand{\Dt}{\tilde D}

\newcommand{\xR}{x^r}

\newcommand{\cotrainSS}{\textsc{S\textsuperscript{2}GAN-CO}} 
\newcommand{\cotrainSSS}{\textsc{S\textsuperscript{3}GAN-CO}}
\newcommand{\tranSS}{\textsc{S\textsuperscript{2}GAN}}
\newcommand{\tranSSS}{\textsc{S\textsuperscript{3}GAN}}
\newcommand{\tranC}{\textsc{Clustering}}

\newcommand{\slabels}{\textsc{Single label}}

\newcommand{\rlabels}{\textsc{Random label}}

\newcommand{\biggan}{\textsc{BigGAN}}

\icmltitlerunning{
High-Fidelity Image Generation With Fewer Labels}

\begin{document}

\twocolumn[
\icmltitle{
High-Fidelity Image Generation With Fewer Labels}

\icmlsetsymbol{equal}{*}

\begin{icmlauthorlist}
\icmlauthor{Mario Lucic}{equal,brain}
\icmlauthor{Michael Tschannen}{equal,eth}
\icmlauthor{Marvin Ritter}{equal,brain}
\icmlauthor{Xiaohua Zhai}{brain}
\icmlauthor{Olivier Bachem}{brain}
\icmlauthor{Sylvain Gelly}{brain}
\end{icmlauthorlist}

\icmlaffiliation{brain}{Google Research, Brain Team}
\icmlaffiliation{eth}{ETH Zurich}
\icmlcorrespondingauthor{Mario Lucic}{lucic@google.com}
\icmlcorrespondingauthor{Michael Tschannen}{mi.tschannen@gmail.com}
\icmlcorrespondingauthor{Marvin Ritter}{marvinritter@google.com}
\icmlkeywords{Natural image generation,self-supervision}
\vskip 0.3in
]

\printAffiliationsAndNotice{\icmlEqualContribution}

\begin{abstract}
Deep generative models are becoming a cornerstone of modern machine learning. Recent work on conditional generative adversarial networks has shown that learning complex, high-dimensional distributions over natural images is within reach. While the latest models are able to generate high-fidelity, diverse natural images at high resolution, they rely on a vast quantity of labeled data. In this work we demonstrate how one can benefit from recent work on self- and semi-supervised learning to outperform the state of the art on both unsupervised ImageNet synthesis, as well as in the conditional setting. In particular, the proposed approach is able to match the sample quality (as measured by FID) of the current state-of-the-art conditional model BigGAN on ImageNet \emph{using only 10\% of the labels} and outperform it using 20\% of the labels.
\end{abstract}

\vspace{-5mm}
\section{Introduction}

\looseness-1 Deep generative models have received a great deal of attention due to their power to learn complex high-dimensional distributions, such as distributions over natural images~\citep{van2016conditional, dinh2016density, brock2018large}, videos~\citep{kalchbrenner2017video}, and audio~\citep{van2016wavenet}. Recent progress was driven by scalable training of large-scale models \cite{brock2018large, menick2018generating}, architectural modifications~\citep{zhang2018self, chen2018selfmodulation,karras2018style}, and normalization techniques~\citep{miyato2018spectral}.

\begin{figure}[t]
\centering
\includegraphics[width=0.45\textwidth]{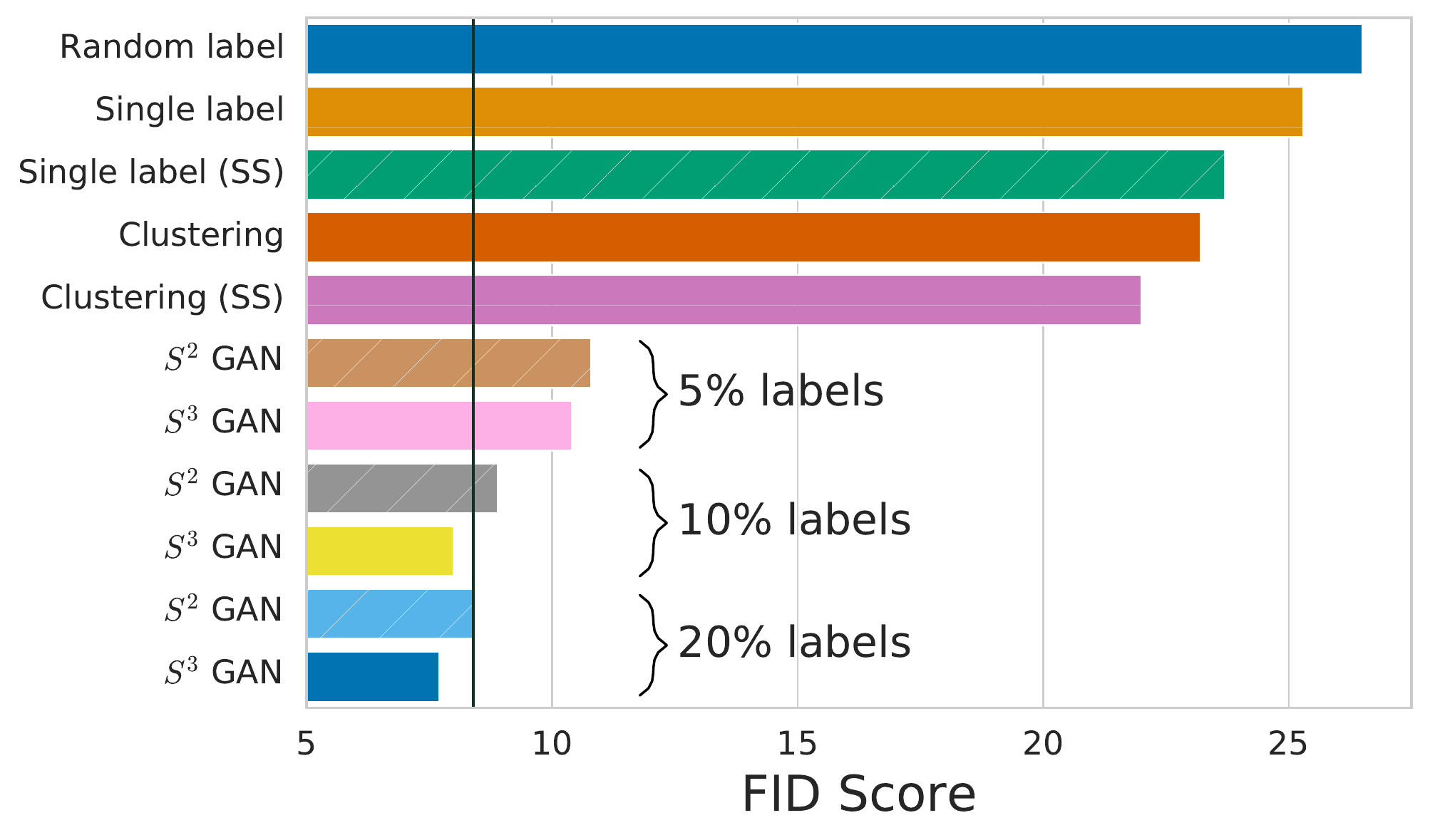}\vspace{-4mm}
\caption{Median FID of the baselines and the proposed method. The vertical line indicates the baseline (\biggan{}) which uses all the labeled data. The proposed method (\tranSSS) is able to match the state-of-the-art while using only $10\%$ of the labeled data and outperform it with 20$\%$.\vspace{-4mm}}
\end{figure}

Currently, high-fidelity natural image generation hinges upon having access to vast quantities of labeled data. The labels induce rich side information into the training process which effectively decomposes the extremely challenging image generation task into semantically meaningful sub-tasks. 

However, this dependence on vast quantities of labeled data is at odds with the fact that most data is unlabeled, and labeling itself is often costly and error-prone. Despite the recent progress on unsupervised image generation, the gap between conditional and unsupervised models in terms of sample quality is significant. 

In this work, we take a significant step towards closing the gap between conditional and unsupervised generation of high-fidelity images using generative adversarial networks (GANs). We leverage two simple yet powerful concepts:
\begin{enumerate}[itemsep=0pt,topsep=-4pt,parsep=0pt]
\item[(i)] Self-supervised learning: A semantic feature extractor for the training data can be learned via self-supervision, and the resulting feature representation can then be employed to guide the GAN training process.
\item[(ii)] Semi-supervised learning: Labels for the entire training set can be inferred from a small subset of labeled training images and the inferred labels can be used as conditional information for GAN training.
\end{enumerate}

\textbf{Our contributions}\quad In this work, we
\begin{enumerate}[itemsep=0pt,topsep=-5pt,parsep=1pt,leftmargin=5mm]
\item propose and study various approaches to reduce or fully omit ground-truth label information for natural image generation tasks,
\item achieve a new state of the art (SOTA) in unsupervised generation on \imagenet{}, match the SOTA on $128\times128$ \imagenet{} using only 10\% of the labels, and set a new SOTA (measured by FID) using 20\% of the labels, and
\item open-source all the code used for the experiments at \href{https://github.com/google/compare_gan}{\texttt{github.com\slash google\slash compare\_gan}}.
\end{enumerate}

\section{Background and related work}
\paragraph{High-fidelity GANs on \imagenet{}} Besides \biggan{} \cite{brock2018large} only a few prior methods have managed to scale GANs to ImageNet, most of them relying on class-conditional generation using labels. One of the earliest attempts are GANs with auxiliary classifier (AC-GANs) \cite{odena2017conditional} which feed one-hot encoded label information with the latent code to the generator and equip the discriminator with an auxiliary head predicting the image class in addition to whether the input is real or fake. More recent approaches rely on a label projection layer in the discriminator, essentially resulting in per-class real/fake classification \cite{miyato2018cgans}, and  self-attention in the generator \cite{zhang2018self}. Both methods use modulated batch normalization \cite{de2017modulating} to provide label information to the generator. On the unsupervised side, \citet{chen2019self} showed that auxiliary rotation loss added to the discriminator has a stabilizing effect on the training. Finally, appropriate gradient regularization enables scaling MMD-GANs to ImageNet without using labels~\cite{arbel2018gradient}.

\textbf{Semi-supervised GANs} \quad Several recent works leveraged GANs for semi-supervised learning of classifiers. Both \citet{salimans2016improved} and \citet{odena2016semi} train a discriminator that classifies its input into $K+1$ classes: $K$ image classes for real images, and one class for generated images. Similarly, \citet{springenberg2015unsupervised} extends the standard GAN objective to $K$ classes. This approach was also considered by \citet{li2017triple} where separate discriminator and classifier models are applied. Other approaches incorporate inference models to predict missing labels \citep{deng2017structured} or harness joint distribution (of labels and data) matching for semi-supervised learning \citep{gan2017triangle}. Up to our knowledge, improvements in sample quality through partial label information are only reported in \citet{li2017triple, deng2017structured, sricharan2017semi}, all of which consider only low-resolution data sets from a restricted domain.

\textbf{Self-supervised learning}\quad Self-supervised learning methods employ a label-free auxiliary task to learn a semantic feature representation of the data. This approach was successfully applied to different data modalities, such as images ~\citep{doersch2015unsupervised, caron2018deep}, video~\citep{agrawal2015learning,lee2017unsupervised}, and robotics~\citep{jang2018grasp2vec,pinto2016supersizing}. The current state-of-the-art method on \imagenet{} is due to~\citet{gidaris2018unsupervised} who proposed predicting the rotation angle of rotated training images as an auxiliary task. This simple self-supervision approach yields representations which are useful for downstream image classification tasks. Other forms of self-supervision include predicting relative locations of disjoint image patches of a given image ~\citep{doersch2015unsupervised, mundhenk2018improvements} or estimating the permutation of randomly swapped image patches on a regular grid~\citep{noroozi2016unsupervised}. A study on self-supervised learning with modern neural architectures is provided in~\citet{kolesnikov2019revisiting}.

\section{Reducing the appetite for labeled data}
In a nutshell, instead of providing hand-annotated ground truth labels for real images to the discriminator, we will provide inferred ones. To obtain these labels we will make use of recent advancements in self- and semi-supervised learning. We propose and study several different methods with different degrees of computational and conceptual complexity. We emphasize that \emph{our work focuses on using few labels to improve the quality of the generative model}, rather than training a powerful classifier from a few labels as extensively studied in prior work on semi-supervised GANs. 

Before introducing these methods in detail, we discuss how label information is used in state-of-the-art GANs. The following exposition assumes familiarity with the basics of the GAN framework \cite{goodfellow2014generative}. 

\begin{figure}[t]
\centering
\includegraphics[width=0.48\textwidth]{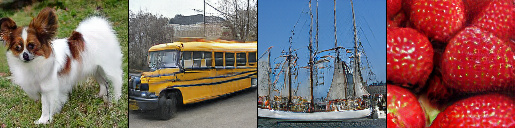}
\includegraphics[width=0.48\textwidth]{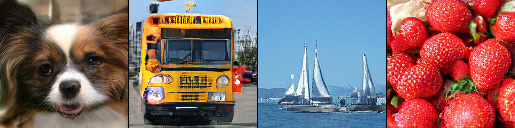}
\caption{Top row: $128 \times 128$ samples from our implementation of the fully supervised current SOTA model \biggan{}. Bottom row: Samples form the proposed \tranSSS{} which matches \biggan{} in terms of FID and IS using only $10\%$ of the ground-truth labels.} 
\end{figure}

\textbf{Incorporating the labels}\quad To provide the label information to the discriminator we employ a linear projection layer as proposed by \citet{miyato2018cgans}. To make the exposition self-contained, we will briefly recall the main ideas. In a ``vanilla" (unconditional) GAN, the discriminator $D$ learns to predict whether the image $x$ at its input is real or generated by the generator $G$. We decompose the discriminator into a learned discriminator representation, $\Dt$, which is fed into a linear classifier, $\cRF$, i.e., the discriminator is given by $\cRF(\Dt(x))$. In the \emph{projection discriminator}, one learns an embedding for each class of the same dimension as the representation $\Dt(x)$. Then, for a given image, label input $x, y$ the decision on whether the sample is real or generated is based on two components: (a) on whether the representation $\Dt(x)$ itself is consistent with the real data, and (b) on whether the representation $\Dt(x)$ is consistent with the real data \emph{from class $y$}. More formally, the discriminator takes the form $D(x,y) = \cRF(\Dt(x)) + P(\Dt(x), y)$, where 
$P(\tilde x, y) = {\tilde x}^\top W y$ is a linear projection layer with learned weight matrix $W$ applied to a feature vector $\tilde x$ and the one-hot encoded label $y$ as an input. 
As for the generator, the label information $y$ is incorporated through class-conditional BatchNorm \cite{dumoulin2017learned, de2017modulating}. The conditional GAN with projection discriminator is illustrated in Figure~\ref{fig:projdisc}.

We proceed with describing the proposed pre-trained and co-training approaches to infer labels for GAN training in Sections~\ref{sec:transfermethods} and~\ref{sec:directmethods}, respectively.

\newcommand{\sw}{0.35\textwidth}
\newcommand{\deflabels}{
\node[] at (0.22, 0.73) {$G$};
 \node[] at (-0.01, 0.8) {$y_\text{f}$};
 \node[] at (0, 0.67) {$z$};
 \node[] at (0.67, 0.48) {$\tilde D$};
 \node[] at (0.48, 0.85) {$x_\text{f}$};
 \node[] at (0.36, 0.12) {$x_\text{r}$};
 \node[] at (0.97, 0.21) {$y_\text{f}$};
 \node[] at (0.885, 0.54) {\footnotesize$\cRF$};
 \node[] at (0.885, 0.42) {\footnotesize$P$};
}

\begin{figure}[t]
\centering
\begin{tikzpicture}
\tikzstyle{every node}=[font=\small]
     \node[anchor=south west,inner sep=0] (imagesp) at (0,0) 
{\includegraphics[width=\sw]{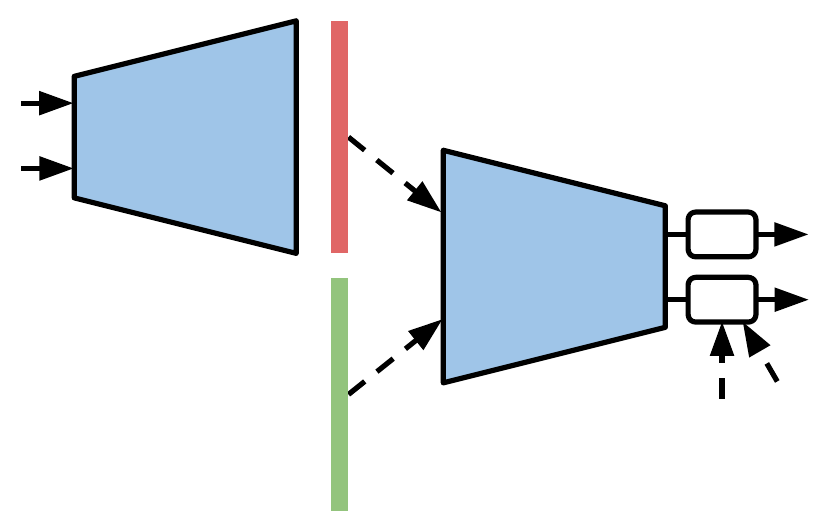}};
     \begin{scope}[x={(imagesp.south east)},y={(imagesp.north west)}]
        \deflabels
         \node[] at (0.89, 0.18) {$y_\text{r}$};
     \end{scope}
\end{tikzpicture}
\caption{\label{fig:projdisc}Conditional GAN with projection discriminator. The discriminator tries to predict from the representation $\Dt$ whether a real image $x_\text{r}$ (with label $y_\text{r}$) or a generated image $x_\text{f}$ (with label $y_\text{f}$) is at its input, by combining an unconditional classifier $\cRF$ and a class-conditional classifier implemented through the projection layer $P$. This form of conditioning is used in \biggan{}. Outward-pointing arrows feed into losses. \label{fig:biggan_illustration}}
\end{figure}

\subsection{Pre-trained approaches} \label{sec:transfermethods}
\begin{figure}[t]
\centering
\begin{tikzpicture}
     \node[anchor=south west,inner sep=0] (image) at (0,0) 
{\includegraphics[width=\sw]{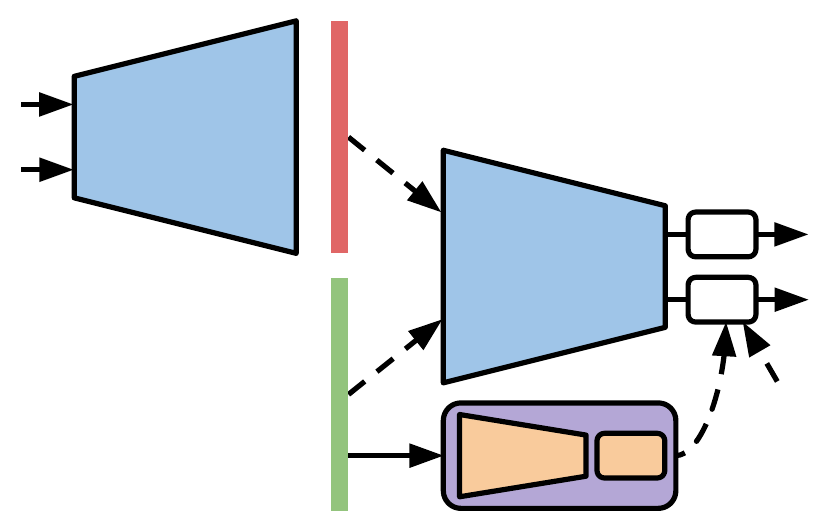}};
     \begin{scope}[x={(image.south east)},y={(image.north west)}]
         \deflabels
         \node[] at (0.63, 0.11) {$F$};
         \node[] at (0.78, 0.12) {\footnotesize$\cCL$};
     \end{scope}
\end{tikzpicture}
\caption{\tranC{}: Unsupervised approach based on clustering the representations obtained by solving a self-supervised task. $F$ corresponds to the feature extractor learned via self-supervision and $\cCL$ is the cluster assignment function. After learning $F$ and $\cCL$ on the real training images in the pre-training step, we proceed with conditional GAN training by inferring the labels as $\yCL = \cCL(F(x))$. \label{fig:unsupervised_illustration}}
\end{figure}

\paragraph{Unsupervised clustering-based method}
We first learn a representation of the real training data using a state-of-the-art self-supervised approach~\citep{gidaris2018unsupervised, kolesnikov2019revisiting}, perform clustering on this representation, and use the cluster assignments as a replacement for labels. Following~\citet{gidaris2018unsupervised} we learn the feature extractor $F$ (typically a convolutional neural network) by minimizing the following \emph{self-supervision loss}
\begin{equation}
\Ls_\text{R} = -\frac{1}{|\mathcal{R}|} \sum_{r \in \mathcal{R}} \E_{x \sim \pdata(x)}[\log p(\cRot(F(x^r))=r)], \label{eq:rotlossF}
\end{equation}
where $\mathcal{R}$ is the set of the $4$ rotation degrees $\{\ang{0}, \ang{90}, \ang{180}, \ang{270}\}$, $\xR$ is the image $x$ rotated by $r$, and $\cRot$ is a linear classifier predicting the rotation degree $r$. After learning the feature extractor $F$, we apply mini batch $k$-Means clustering \citep{sculley2010web} on the representations of the training images. Finally, given the cluster assignment function $\yCL = \cCL(F(x))$ we train the GAN using the hinge loss, alternatively minimizing the discriminator loss $\Ls_D$ and generator loss~$\Ls_G$, namely
\begin{align*}
\Ls_D &= -\E_{x \sim \pdata(x)}[\min(0, -1+D(x,\cCL(F(x))))] \nonumber\\
& \quad- \E_{(z,y) \sim \hat p(z,y)}[\min(0,-1-D(G(z,y),y))]\nonumber\\
\Ls_G &= -\E_{(z,y)\sim \hat p(z,y)}[D(G(z,y),y)],
\end{align*}
where $\hat p(z,y)= p(z)\hat p(y)$ is the prior distribution with $p(z)=\mathcal{N}(0,I)$ and $\hat p(y)$ the empirical distribution of the cluster labels $\cCL(F(x))$ over the training set. 
We call this approach \tranC{} and illustrate it in Figure~\ref{fig:unsupervised_illustration}.

\paragraph{Semi-supervised method}
While semi-supervised learning is an active area of research and a large variety of algorithms has been proposed, we follow~\citet{zhai2019s4l} and simply extend the self-supervised approach described in the previous paragraph with a semi-supervised loss. This ensures that the two approaches are comparable in terms of model capacity and computational cost. Assuming we are provided with labels for a subset of the training data, we attempt to learn a good feature representation via self-supervision and simultaneously train a good linear classifier on the so-obtained representation (using the provided labels).\footnote{Note that an even simpler approach would be to first learn the representation via self-supervision and \emph{subsequently} the linear classifier, but we observed that learning the representation and classifier simultaneously leads to better results.} More formally, we minimize the loss
\begin{align} 
\Ls_\text{S\textsuperscript{2}L} &= - \frac{1}{|\mathcal{R}|} \sum_{r \in \mathcal{R}} \Big\{\E_{x \sim \pdata(x)}[\log p(\cRot(F(\xR))=r)] \nonumber \\
&\quad+  \gamma \E_{(x, y) \sim \pdata(x,y)}[\log p(\cSSL(F(\xR))=y)]\Big\}, \label{eq:ssllloss}
\end{align}
where $\cRot$ and $\cSSL$ are linear classifiers predicting the rotation angle $r$ and the label $y$, respectively, and $\gamma>0$ balances the loss terms. The first term in~\eqref{eq:ssllloss} corresponds to the self-supervision loss from~\eqref{eq:rotlossF} and the second term to a (semi-supervised) cross-entropy loss. During training, the latter expectation is replaced by the empirical average over the subset of labeled training examples, whereas the former is set to the empirical average over the entire training set (this convention is followed throughout the paper). After we obtain $F$ and $\cSSL$ we proceed with GAN training where we label the real images as $\ySSL = \cSSL(F(x))$. In particular, we alternatively minimize the same generator and discriminator losses as for \tranC{} except that we use $\cSSL$ and $F$ obtained by minimizing \eqref{eq:ssllloss}:
\begin{align*}
\Ls_D &= - \E_{x \sim \pdata(x)}[\min(0, -1+D(x,\cSSL(F(x))))] \\
      & \quad- \E_{(z,y) \sim p(z,y)}[\min(0,-1-D(G(z,y),y))] \\
\Ls_G &= -\E_{(z,y)\sim p(z,y)}[D(G(z,y),y)],
\end{align*}
where $p(z,y)= p(z)p(y)$ with $p(z)=\mathcal{N}(0,I)$ and $p(y)$ uniform categorical. We use the abbreviation \tranSS{} for this method.

\subsection{Co-training approach} \label{sec:directmethods}
The main drawback of the transfer-based methods is that one needs to train a feature extractor $F$ via self-supervision and learn an inference mechanism for the labels (linear classifier or clustering). In what follows we detail co-training approaches that avoid this two-step procedure and learn to infer label information during GAN training.

\begin{figure}[t]
\centering
\begin{tikzpicture}
     \node[anchor=south west,inner sep=0] (image) at (0,0) 
{\includegraphics[width=\sw]{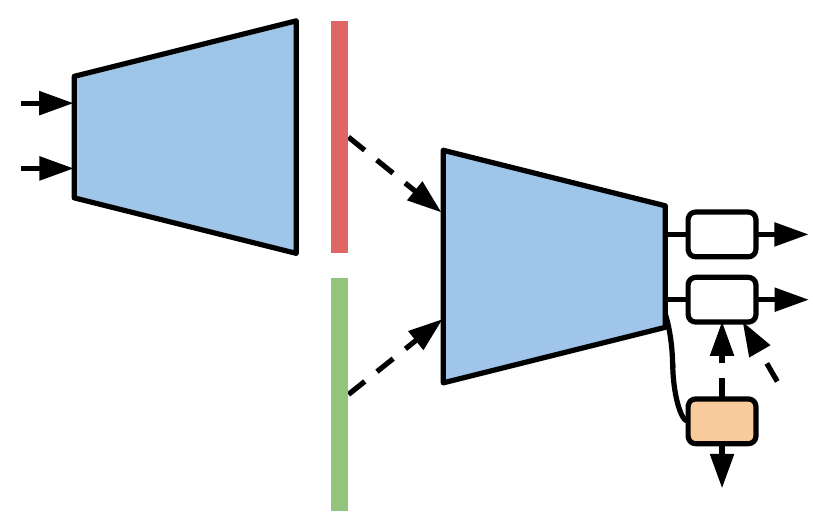}};
     \begin{scope}[x={(image.south east)},y={(image.north west)}]
         \deflabels
         \node[] at (0.89, 0.18) {$\cCT$};
     \end{scope}
\end{tikzpicture}
\caption{\cotrainSS{}: During GAN training we learn an auxiliary classifier $\cCT$ on the discriminator representation $\Dt$, based on the labeled real examples, to predict labels for the unlabeled ones. This avoids training a feature extractor $F$ and classifier $\cSSL$ prior to GAN training as in \tranSS{}. \label{fig:cotrained}} 
\end{figure}
\textbf{Unsupervised method}\quad
We consider two approaches. In the first one, we completely remove the labels by simply labeling all real and generated examples with the same label\footnote{Note that this is not necessarily equivalent to replacing class-conditional BatchNorm with standard (unconditional) BatchNorm as the variant of conditional BatchNorm used in this paper also uses chunks of the latent code as input;  besides the label information.} and removing the projection layer from the discriminator, i.e., we set $D(x) = \cRF(\Dt(x))$. We use the abbreviation \slabels{} for this method. For the second approach we assign random labels to (unlabeled) real images. While the labels for the real images do not provide any useful signal to the discriminator, the sampled labels could potentially help the generator by providing additional randomness with different statistics than $z$, as well as additional trainable parameters due to the embedding matrices in class-conditional BatchNorm. Furthermore, the labels for the fake data could facilitate the discrimination as they provide side information about the fake images to the discriminator. We term this method \rlabels{}.

\textbf{Semi-supervised method}\quad When labels are available for a subset of the real data, we train an auxiliary linear classifier $\cCT$ directly on the feature representation $\Dt$ of the discriminator, \emph{during GAN training}, and use it to predict labels for the unlabeled real images. In this case the discriminator loss takes the form
\begin{align}
\Ls_D = &- \E_{(x, y) \sim \pdata(x,y)}[\min(0, -1+D(x, y))] \nonumber \\
& - \lambda \E_{(x, y) \sim \pdata(x,y)}[\log p(\cCT(\Dt(x))=y)] \nonumber \\
& - \E_{x \sim \pdata(x)}[\min(0, -1+D(x, \cCT(\Dt(x))))] \nonumber \\
& - \E_{(z,y) \sim p(z,y)}[\min(0,-1-D(G(z,y),y))], \label{eq:cotrain}
\end{align}
where the first term corresponds to standard conditional training on ($k\%$) labeled real images, the second term is the cross-entropy loss (with weight $\lambda>0$) for the auxiliary classifier $\cCT$ on the labeled real images, the third term is an unsupervised discriminator loss where the labels for the unlabeled real images are predicted by $\cCT$, and the last term is the standard conditional discriminator loss on the generated data. We use the abbreviation \cotrainSS{} for this method. See Figure~\ref{fig:cotrained} for an illustration.

\subsection{Self-supervision during GAN training}\label{sec:selfsup}
\begin{figure}[t!]
\centering
\begin{tikzpicture}
     \node[anchor=south west,inner sep=0] (image) at (0,0) 
{\includegraphics[width=\sw]{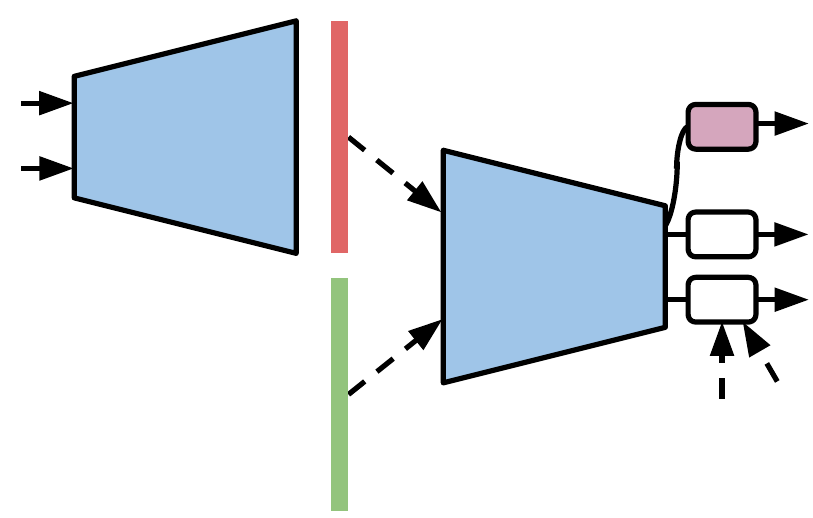}};
     \begin{scope}[x={(image.south east)},y={(image.north west)}]
         \deflabels
         \node[] at (0.89, 0.18) {$y_\text{r}$};
         \node[] at (0.89, 0.75) {\footnotesize$\cRot$};
     \end{scope}
\end{tikzpicture}
\caption{Self-supervision by rotation-prediction during GAN training. Additionally to predicting whether the images at its input are real or generated, the discriminator is trained to predict rotations of both rotated real and fake images via an auxiliary linear classifier $\cRot$. This approach was successfully applied by~\citet{chen2019self} to stabilize GAN training. Here we combine it with our pre-trained and co-training approaches, replacing the ground truth labels $y_\text{r}$ with predicted ones.\label{fig:ssgan}}
\vspace{-0.2cm}
\end{figure}
So far we leveraged self-supervision to either craft good feature representations, or to learn a semi-supervised model (cf. Section~\ref{sec:transfermethods}). However, given that the discriminator itself is just a classifier, one may benefit from augmenting this classifier with an auxiliary task---namely self-supervision through rotation prediction. This approach was already explored in~\citet{chen2019self}, where it was observed to stabilize GAN training. Here we want to assess its impact when combined with the methods introduced in Sections~\ref{sec:transfermethods} and~\ref{sec:directmethods}. To this end, similarly to the training of $F$ in \eqref{eq:rotlossF} and \eqref{eq:ssllloss}, we train an additional linear classifier $\cRot$ on the discriminator feature representation $\Dt$ to predict rotations $r \in \mathcal R$ of the rotated real images $\xR$ and rotated fake images $G(z,y)^r$. The corresponding loss terms added to the discriminator and generator losses are
\begin{equation}\label{eq:ssganD}
-\frac{\beta}{|\mathcal{R}|} \sum_{r \in \mathcal{R}} \E_{x \sim \pdata(x)}[\log p(\cRot(\Dt(\xR)=r)]
\end{equation}
and
\begin{equation}\label{eq:ssganG}
-\frac{\alpha}{|\mathcal{R}|} \E_{(z,y) \sim p(z,y)}[\log p(\cRot(\Dt(G(z,y)^r)=r)],
\end{equation}
respectively, where $\alpha, \beta>0$ are weights to balance the loss terms. This approach is illustrated in Figure~\ref{fig:ssgan}.

\section{Experimental setup}\label{exp:setup}

\paragraph{Architecture and hyperparameters} 
GANs are notoriously unstable to train and their performance strongly depends on the capacity of the neural architecture, optimization hyperparameters, and appropriate regularization~\cite{lucic2018,kurach2018gan}. We implemented the conditional \biggan{} architecture \cite{brock2018large} which achieves state-of-the-art results on ImageNet.\footnote{We dissected the model checkpoints released by \citet{brock2018large} to obtain exact counts of trainable parameters and their dimensions, and match them to \emph{byte} level (cf. Tables~\ref{tab:discriminator_details} and ~\ref{tab:generator_details} in Appendix~\ref{sec:arch_details}). We want to emphasize that at this point this methodology is \emph{bleeding-edge} and successful state-of-the-art methods require careful architecture-level tuning. To foster reproducibility we meticulously detail this architecture at tensor-level detail in Appendix~\ref{sec:arch_details} and open-source our code at \url{https://github.com/google/compare_gan}.}  We use exactly the same optimization hyper-parameters as \citet{brock2018large}. Specifically, we employ the Adam optimizer with the learning rates $5\cdot10^{-5}$ for the generator and $2\cdot10^{-4}$ for the discriminator ($\beta_1=0$, $\beta_2=0.999$). We train for 250k generator steps with 2 discriminator steps before each generator step. The batch size was fixed to 2048, and we use a latent code $z$ with $120$ dimensions. We employ spectral normalization in both generator and discriminator. In contrast to \biggan{}, we do not apply orthogonal regularization as this was observed to only marginally improve sample quality (cf. Table 1 in \citet{brock2018large}) and we do not use the truncation trick.

\textbf{Datasets}\quad We focus primarily on \imagenet{}, the largest and most diverse image data set commonly used to evaluate GANs. \imagenet{} contains $1.3$M training images and $50$k test images, each corresponding to one of 1k object classes. We resize the images to $128\times128\times3$ as done in~\citet{miyato2018cgans} and ~\citet{zhang2018self}. Partially labeled data sets for the semi-supervised approaches are obtained by randomly selecting $k\%$ of the samples from each class.

\textbf{Evaluation metrics} \quad
We use the Fr{\'e}chet Inception Distance (FID)~\citep{heusel2017gans} and Inception Score~\citep{salimans2016improved} to evaluate the quality of the generated samples. To compute the FID, the real data and generated samples are first embedded in a specific layer of a pre-trained Inception network. Then, a multivariate Gaussian is fit to the data and the distance computed as
$\FID(x, g) = ||\mu_x -\mu_g||_2^2 + \Tr(\Sigma_x + \Sigma_g - 2(\Sigma_x\Sigma_g)^\frac12)$,
where $\mu$ and $\Sigma$ denote the empirical mean and covariance, and subscripts $x$ and $g$ denote the real and generated data respectively. FID was shown to be sensitive to both the addition of spurious modes and to mode dropping~\citep{sajjadi2018assessing,lucic2018}. Inception Score posits that conditional label distribution of samples containing meaningful objects should have low entropy, and the variability of the samples should be
high leading to the following formulation: $\text{IS} = \exp(\E_{x \sim Q}[d_{KL}(p(y
\mid x), p(y))])$. Although it has some flaws~\citep{barratt2018note}, we report it to enable comparison with existing methods. Following~\citet{brock2018large}, the FID is computed using the 50k \imagenet{} testing images and 50k randomly sampled fake images, and the IS is computed from 50k randomly sampled fake images. All metrics are computed for 5 different randomly sampled sets of fake images and are then averaged.

\begin{table}[t]
  \centering
  \caption{A short summary of the analyzed methods. The detailed descriptions of pre-training and co-trained approaches can be found in Sections~\ref{sec:transfermethods} and \ref{sec:directmethods}, respectively. Self-supervision during GAN training is described in Section~\ref{sec:selfsup}.\vspace{0.2cm}}
\begin{tabular}{ll}
\toprule
  \textsc{Method} & \textsc{Description} \\\midrule
  \biggan         & Conditional \citep{brock2018large} \\\midrule
  \slabels        & Co-training: Single label\\
  \rlabels        & Co-training: Random labels\\
  \tranC{}        & Pre-trained: Clustering\\\midrule
  \biggan-$k\%$    & \biggan{} using only $k\%$ labeled data \\
  \cotrainSS{}    & Co-training: Semi-supervised\\
  \tranSS{}       & Pre-trained: Semi-supervised\\\midrule
  \tranSSS{}      & \tranSS{} with self-supervision\\
  \cotrainSSS{}   & \cotrainSS{} with self-supervision\\
  \bottomrule
\end{tabular}
\label{tab:methods}
\end{table}

\textbf{Methods}\quad
We conduct an extensive comparison of methods detailed in Table~\ref{tab:methods}, namely: Unmodified \biggan, the unsupervised methods \slabels{}, \rlabels{}, \tranC{}, and the semi-supervised methods \tranSS{} and \cotrainSS{}. In all \cotrainSS{} experiments we use soft labels, i.e., the soft-max output of $\cCT$ instead of one-hot encoded hard estimates, as we observed in preliminary experiments that this stabilizes training. For \tranSS{} we use hard labels by default, but investigate the effect of soft labels in separate experiments. For all semi-supervised methods we have access only to $k\%$ of the ground truth labels where $k\in\{5, 10, 20\}$. As an additional baseline, we retain $k\%$ labeled real images and discard all unlabeled real images, then using the remaining labeled images to train \biggan{} (the resulting model is designated by \biggan-$k\%$). Finally, we explore the effect of self-supervision during GAN training on the unsupervised and semi-supervised methods. 

We train every model three times with a different random seed and report the median FID and the median IS. With the exception of the \slabels{} and \biggan-$k\%$, the standard deviation of the mean across three runs is very low. We therefore defer tables with the mean FID and IS values and standard deviations to Appendix~\ref{app:meanstd}. All models are trained on 128 cores of a Google TPU v3 Pod with BatchNorm statistics synchronized across cores.

\emph{Unsupervised approaches} For~\tranC{} we simply used the best available self-supervised rotation model from~\citet{kolesnikov2019revisiting}. The number of clusters for \tranC{} is selected from the set $\{50, 100, 200, 500, 1000\}$. The other unsupervised approaches do not have hyper-parameters.

\emph{Pre-trained and co-training approaches}
We employ the wide ResNet-50 v2 architecture with widening factor 16 ~\cite{zagoruyko2016wide} for the feature extractor $F$ in the pre-trained approaches described in Section~\ref{sec:transfermethods}. 

We optimize the loss in~\eqref{eq:ssllloss} using SGD for 65 epochs. The batch size is set to $2048$, composed of $B$ unlabeled examples and $2048-B$ labeled examples. Following the recommendations from ~\citet{goyal2017accurate} for training with large batch size, we (i) set the learning rate to $0.1\frac {B} {256}$, and (ii) use linear learning rate warm-up during the initial 5 epochs. The learning rate is decayed twice with a factor of 10 at epoch 45 and epoch 55. The parameter $\gamma$ in \eqref{eq:ssllloss} is set to $0.5$ and the number of unlabeled examples per batch $B$ is 1536. The parameters $\gamma$ and $B$ are tuned on $0.1\%$ labeled examples held out from the training set, the search space is $\{0.1, 0.5, 1.0\} \times \{1024, 1536, 1792\}$. The accuracy of the so-obtained classifier $\cSSL(F(x))$ on the \imagenet{} validation set is reported in Table~\ref{tab:semi_self_supervision}. The parameter $\lambda$ in the loss used for \cotrainSS{} in \eqref{eq:cotrain} is selected form the set $\{0.1, 0.2, 0.4\}$.

\textbf{Self-supervision during GAN training}\quad For all approaches we use the recommend parameter $\alpha=0.2$ from \cite{chen2019self} in \eqref{eq:ssganG} and do a small sweep for $\beta$ in \eqref{eq:ssganD}. For the values tried ($\{0.25, 0.5, 1.0, 2\}$) we do not see a large effect and use $\beta=0.5$ for $\tranSSS$. For $\cotrainSSS$ we did not repeat the sweep, and instead used $\beta=1.0$.

\section{Results and discussion}

Recall that the main goal of this work is to match (or outperform) the fully supervised \biggan{} in an unsupervised fashion, or with a small subset of labeled data. 
In the following, we discuss the advantages and drawbacks of the analyzed approaches with respect to this goal.

As a baseline, our reimplementation of \biggan{} obtains an FID of 8.4 and IS of 75.0, and hence reproduces the result reported by \citet{brock2018large} in terms of FID. We observed some differences in training dynamics, which we discuss in detail in Section~\ref{sec:otherinsights}. 

\subsection{Unsupervised approaches}
\begin{table}[b]
\centering
\caption{Median FID and IS for the unsupervised approaches (see Table~\ref{tab:unsupervised_fid_is_mean_std} in the appendix for mean and standard deviation).\vspace{0.2cm}}
\input{tables/unsupervised_fid_is_median.tex}
\label{tab:unsupervised}
\end{table}

\begin{figure}[h]
\vspace{0.2cm}
\centering\includegraphics[width=0.45\textwidth]{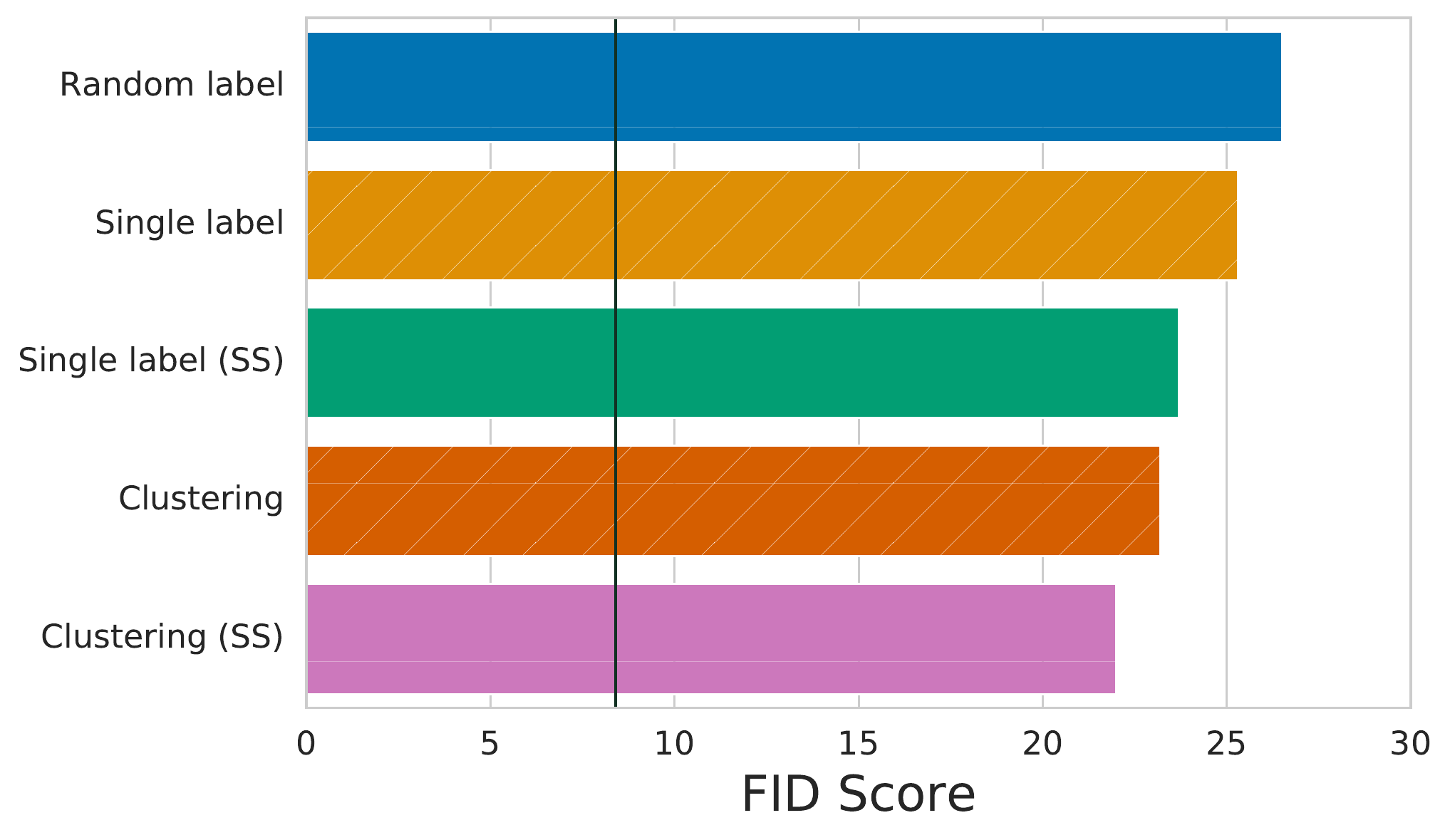}
\caption{Median FID obtained by our unsupervised approaches. The vertical line indicates the median FID of our \biggan{} implementation which uses labels for all training images. While the gap between unsupervised and fully supervised approaches remains significant, using a pre-trained self-supervised representation (\tranC{}) improves the sample quality compared to \slabels{} and \rlabels{}, leading to a new SOTA in unsupervised generation on \imagenet{}.\label{fig:unsupervised}}
\end{figure}

The results for unsupervised approaches are summarized in Figure~\ref{fig:unsupervised} and Table~\ref{tab:unsupervised}. The fully unsupervised \rlabels{} and \slabels{} models both achieve a similar FID of $\sim25$ and IS of $\sim20$. This is a quite considerable gap compared to \biggan{} and indicates that additional supervision is necessary. We note that one of the three \slabels{} models collapsed whereas all three \rlabels{} models trained stably for 250k generator iterations.

Pre-training a semantic representation using self-supervision and clustering the training data on this representation as done by \tranC{} 
reduces the FID by about $10\%$ and increases IS by about $10\%$. These results were obtained for 50 clusters, all other options led to worse results. While this performance is still considerably worse than that of \biggan{} this result is the current SOTA in unsupervised image generation (\citet{chen2019self} report an FID of 33 for unsupervised generation).

Example images from the clustering are shown in Figures~\ref{fig:cluster15samples}, \ref{fig:cluster16samples}, and \ref{fig:cluster42samples} in the supplementary material. The clustering is clearly meaningful and groups similar objects within the same cluster. Furthermore, the objects generated by \tranC{} conditionally on a given cluster index reflect the distribution of the training data belonging to the corresponding cluster. On the other hand, we can clearly observe multiple classes being present in the same cluster. This is to be expected when under-clustering to $50$ clusters. Interestingly, clustering to many more clusters (say $500$) yields results similar to \slabels{}.

\subsection{Semi-supervised approaches}
\textbf{Pre-trained}\quad
The \tranSS{} model where we use the classifier pre-trained with both a self-supervised and semi-supervised loss (cf. Section~\ref{sec:transfermethods}) matches the \biggan{} baseline when $20\%$ of the labels are used and incurs a minor increase in FID when $10\%$ and $5\%$ are used (cf. Table~\ref{tab:semi_self_supervision}). We stress that this is despite the fact that the classifier used to infer the labels has a top-1 accuracy of only $50\%$, $63\%$, and $71\%$ for $5\%$, $10\%$, and $20\%$ labeled data, respectively (cf. Table~\ref{tab:semi_self_supervision}), compared to $100\%$ of the original labels. The results are shown in Table~\ref{tab:transfer_vs_direct} and Figure~\ref{fig:barwinner}, and random samples as well as interpolations can be found in Figures~\ref{fig:s2-20-5}--\ref{fig:s2-20-1} in the supplementary material.
\begin{table}[b]
\centering
\caption{Top-1 and top-5 error rate (\%) on the \imagenet{} validation set of $\cSSL(F(x))$ using both self- and semi-supervised losses as described in Section~\ref{sec:transfermethods}. While the models are clearly not state-of-the-art compared to the fully supervised \imagenet{} classification task, the quality of labels is sufficient to match and in some cases improve the state-of-the-art GAN natural image synthesis.\vspace{0.2cm}}
\input{tables/semi_self_supervision.tex}
\label{tab:semi_self_supervision}
\end{table}

\begin{table}[h]
\caption{Pre-trained vs co-training approaches, and the effect of self-supervision during GAN training (see Table~\ref{tab:transfer_vs_direct_mean_std} in the appendix for mean and standard deviation). While co-training approaches outperform fully unsupervised approaches, they are clearly outperformed by the pre-trained approaches. Self-supervision during GAN training helps in all cases.\vspace{0.2cm}}
\centering
\input{tables/transfer_vs_direct_median.tex}
\label{tab:transfer_vs_direct}
\end{table}

\paragraph{Co-trained}

The results for our co-trained model \cotrainSS{} which trains a linear classifier in semi-supervised fashion on top of the discriminator representation during GAN training (cf. Section~\ref{sec:directmethods}) are shown in Table~\ref{tab:transfer_vs_direct}. It can be seen that \cotrainSS{} outperforms all fully unsupervised approaches for all considered label percentages. While the gap between \cotrainSS{} with $5\%$ labels and \tranC{} in terms of FID is small, \cotrainSS{} has a considerably larger IS. When using $20\%$ labeled training examples \cotrainSS{} obtains an FID of 13.9 and an IS of 49.2, which is remarkably close to \biggan{} and \tranSS{} given the simplicity of the \cotrainSS{} approach. As the the percentage of labels decreases, the gap between \tranSS{} and \cotrainSS{} increases. 

Interestingly, \cotrainSS{} does not seem to train less stably than \tranSS{} approaches even though it is forced to learn the classifier during GAN training. This is particularly remarkable as the \biggan{}-$k\%$ approaches, where we only retain the labeled data for training and discard all unlabeled data, \emph{are very unstable and collapse after 60k to 120k iterations}, for all three random seeds and for both $10\%$ and $20\%$ labeled data.

\subsection{Self-supervision during GAN training}
\begin{figure}[t]
\vspace{0.5cm}
\centering
\includegraphics[width=0.45\textwidth]{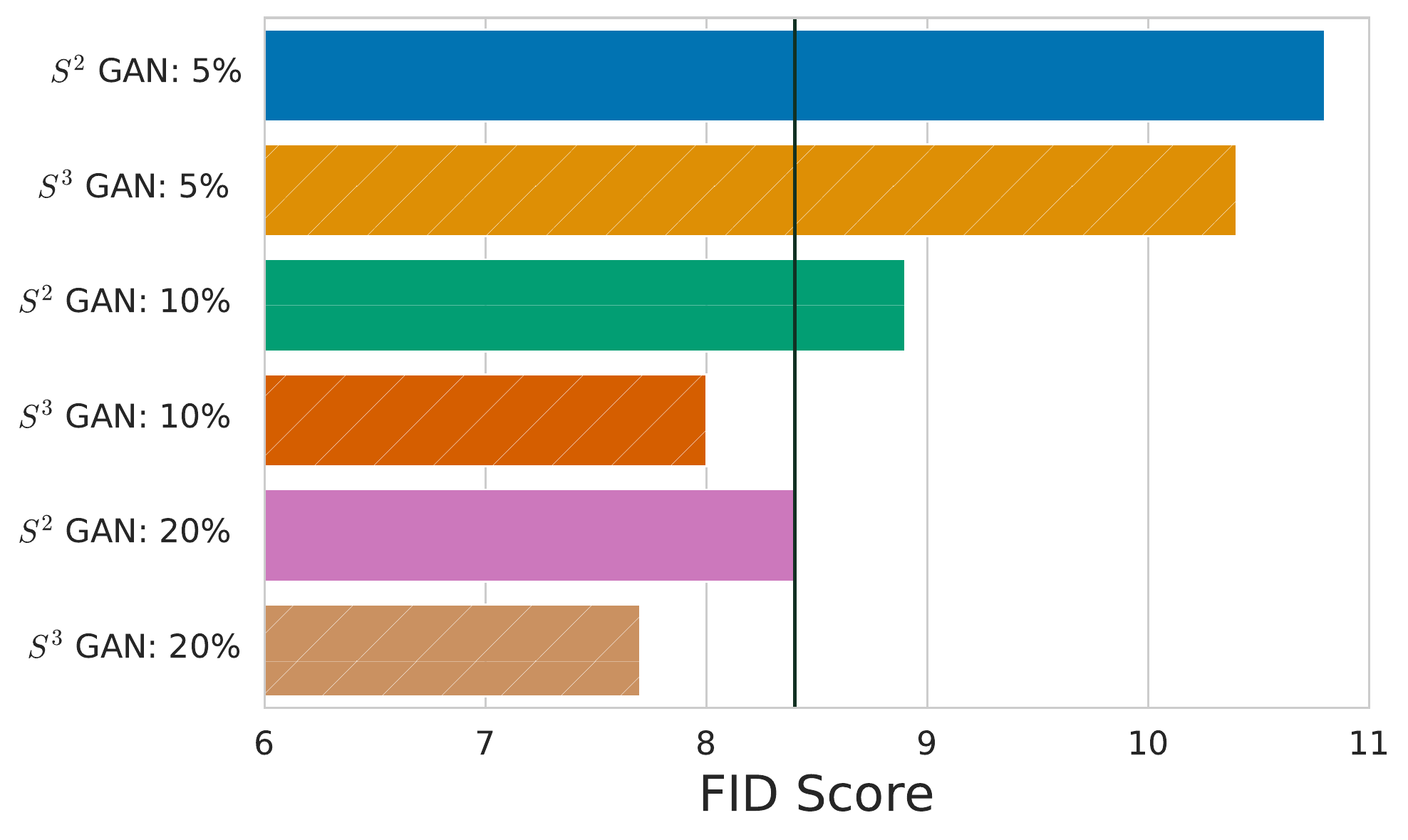}
\caption{\label{fig:barwinner}The vertical line indicates the median FID of our \biggan{} implementation which uses all labeled data. The proposed \tranSSS{} approach is able to match the performance of the state-of-the-art \biggan{} model using $10\%$ of the ground-truth labels and outperforms it using $20\%$.} 
\end{figure}

So far we have seen that the pre-trained semi-supervised approach, namely \tranSS{}, is able to achieve state-of-the-art performance for $20\%$ labeled data. Here we investigate whether self-supervision during GAN training as described in Section~\ref{sec:selfsup} can lead to further improvements. Table~\ref{tab:transfer_vs_direct} and Figure~\ref{fig:barwinner} show the experimental results for \tranSSS{}, namely \tranSS{} coupled with self-supervision in the discriminator.

Self-supervision leads to a reduction in FID and increase in IS across all considered settings. In particular \emph{we can match the state-of-the-art \biggan{} with only $10\%$ of the labels and outperform it using $20\%$ labels, both in terms of FID and IS.}

For \tranSSS{} the improvements due to self-supervision during GAN training in FID are considerable, around $10\%$ in most of the cases. Tuning the parameter $\beta$ of the discriminator self-supervision loss in \eqref{eq:ssganD} did not dramatically increase the benefits of self-supervision during GAN training, at least for the range of values considered. 
As shown in Tables~\ref{tab:unsupervised} and~\ref{tab:transfer_vs_direct}, self-supervision during GAN training (with default parameters $\alpha, \beta$) also leads to improvements by $5$ to $10\%$ for both \cotrainSS{} and \slabels{}. In summary, self-supervision during GAN training with default parameters leads to a stable improvement across all approaches.

\subsection{Other insights} \label{sec:otherinsights}
\textbf{Effect of soft labels}\quad A design choice available to practitioners is whether to use hard labels (i.e., the argmax over the logits), or soft labels (softmax over the logits) for \tranSS{} and \tranSSS{} (recall that we use soft labels by default for \cotrainSS{} and \cotrainSSS{}). Our initial expectation was that soft labels should help when very little labeled data is available, as soft labels carry more information which can potentially be exploited by the projection discriminator. Surprisingly, the results presented in Table~\ref{tab:soft_vs_hard} show clearly that the opposite is true. Our current hypothesis is that this is due to the way labels are incorporated in the projection discriminator, but we do not have empirical evidence yet.

\textbf{Optimization dynamics}\quad \citet{brock2018large} report the FID and IS of the model \emph{just before the collapse}, which can be seen as a form of early stopping. In contrast, we manage to stably train the proposed models for 250k generator iterations. In particular, we also observe stable training for our ``vanilla" \biggan{} implementation. The evolution of the FID and IS as a function of the training steps is shown in Figure~\ref{fig:convergence} in the appendix. At this point we can only speculate about the origin of this difference. We finally note that by tuning the learning rate we obtained slightly different (but still stable) training dynamics in terms of IS, achieving FID~$6.9$ and IS~$98$ for \tranSSS{} with $20\%$ labels.

\begin{table}[t]
\caption{Training with hard (predicted) labels leads to better models than training with soft (predicted) labels (see Table~\ref{tab:soft_vs_hard_mean_std} in the appendix for mean and standard deviation).\vspace{0.2cm}}
\centering
\input{tables/soft_vs_hard_median.tex}
\label{tab:soft_vs_hard}
\end{table}

\textbf{Higher resolution and going below 5\% labels}\quad Training these models at higher resolution becomes computationally harder and it necessitates tuning the learning rate. We trained several \tranSSS{} models at $256 \times 256$ resolution and show the resulting samples in Figures~\ref{fig:s2-20-256-1}--\ref{fig:s2-20-256-2} and interpolations in Figures~\ref{fig:s2-20-256-3}--\ref{fig:s2-20-256-4}.
We also conducted \tranSSS{} experiments in which only $2.5\%$ of the labels are used and observed FID of $13.6$ and IS of $46.3$. This indicates that given a small number of samples one can significantly outperform the unsupervised approaches (c.f. Figure~\ref{fig:unsupervised}).

\section{Conclusion and future Work}
In this work we investigated several avenues to reduce the appetite for labeled data in state-of-the-art GANs. We showed that recent advances in self and semi-supervised learning can be used to achieve a new state of the art, both for unsupervised and supervised natural image synthesis.

We believe that this is a great first step towards the ultimate goal of few-shot high-fidelity image synthesis. There are several important directions for future work: (i) investigating the applicability of these techniques for even larger and more diverse data sets, and (ii) investigating the impact of other self- and semi-supervised approaches on the model quality. (iii) investigating the impact of self-supervision in other deep generative models. Finally, we would like to emphasize that further progress might be hindered by the engineering challenges related to training large-scale generative adversarial networks. To help alleviate this issue and to foster reproducibility, we have open-sourced all the code used for the experiments.

\section*{Acknowledgments}
We would like to thank Ting Chen and Neil Houlsby for fruitful discussions on self-supervision and its application to GANs. We would like to thank Lucas Beyer, Alexander Kolesnikov, and Avital Oliver for helpful discussions on self-supervised semi-supervised learning. We would like to thank Karol Kurach and Marcin Michalski their major contributions the Compare GAN library. We would also like to thank the BigGAN team (Andy Brock, Jeff Donahue, and Karen Simonyan) for their insights into training GANs on TPUs. Finally, we are grateful for the support of members of the Google Brain team in Zurich. This work was partially done while Michael Tschannen was at Google Research.
\bibliography{s3gan}
\bibliographystyle{icml2019}

\appendix
\onecolumn

\section{Additional samples and interpolations}\label{sec:samples}

\begin{figure}[h]
\centering
\makebox[\textwidth][c]{\includegraphics[scale=0.5]{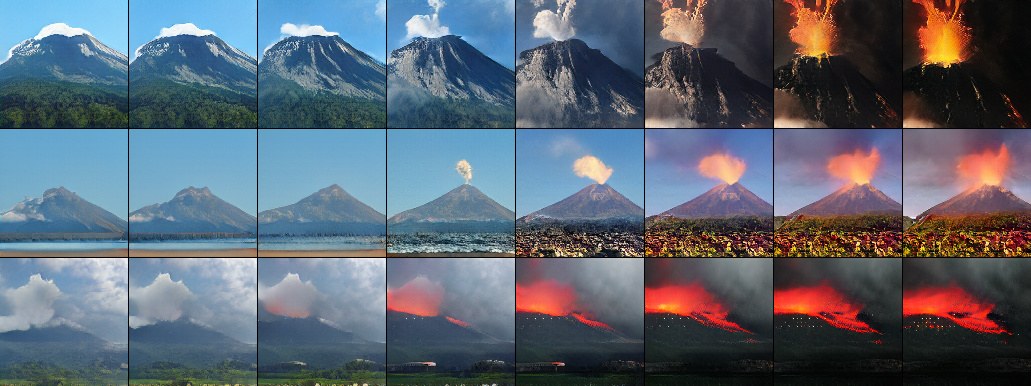}}
\caption{Samples obtained from \tranSSS{} (20\% labels, $128\times128$) when interpolating in the latent space (left to right).\label{fig:s2-20-5}}
\end{figure}
\vspace{3cm}
\begin{figure}[h]
\centering
\makebox[\textwidth][c]{\includegraphics[scale=0.5]{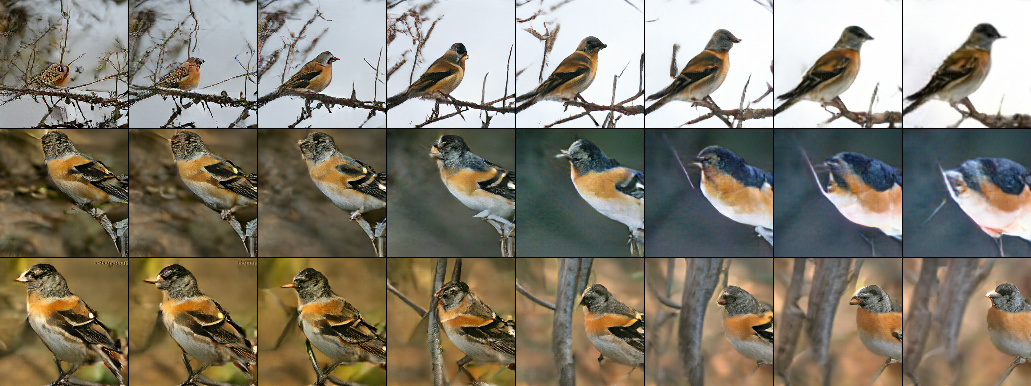}}
\caption{Samples obtained from \tranSSS{} (20\% labels, $128\times128$) when interpolating in the latent space (left to right).\label{fig:s2-20-3}}
\end{figure}

\begin{figure}[h]
\centering
\makebox[\textwidth][c]{\includegraphics[scale=0.5]{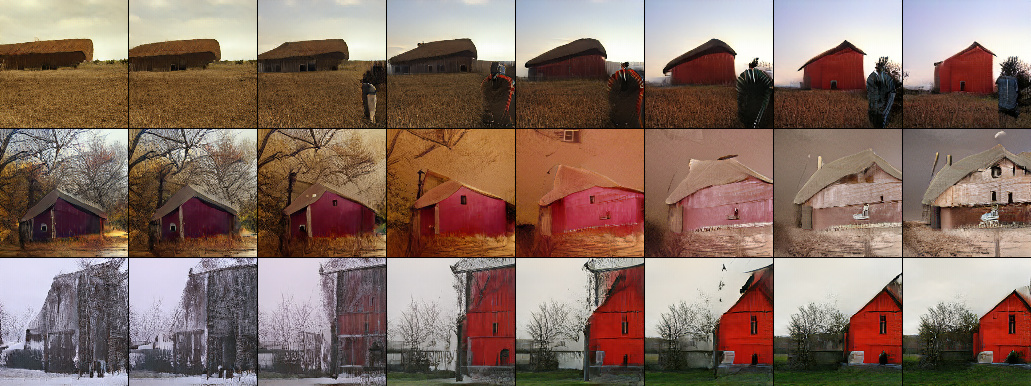}}
\caption{Samples obtained from \tranSSS{} (20\% labels, $128\times128$) when interpolating in the latent space (left to right).\label{fig:s2-20-4}}
\end{figure}

\begin{figure}[h]
\centering
\makebox[\textwidth][c]{\includegraphics[scale=1.4]{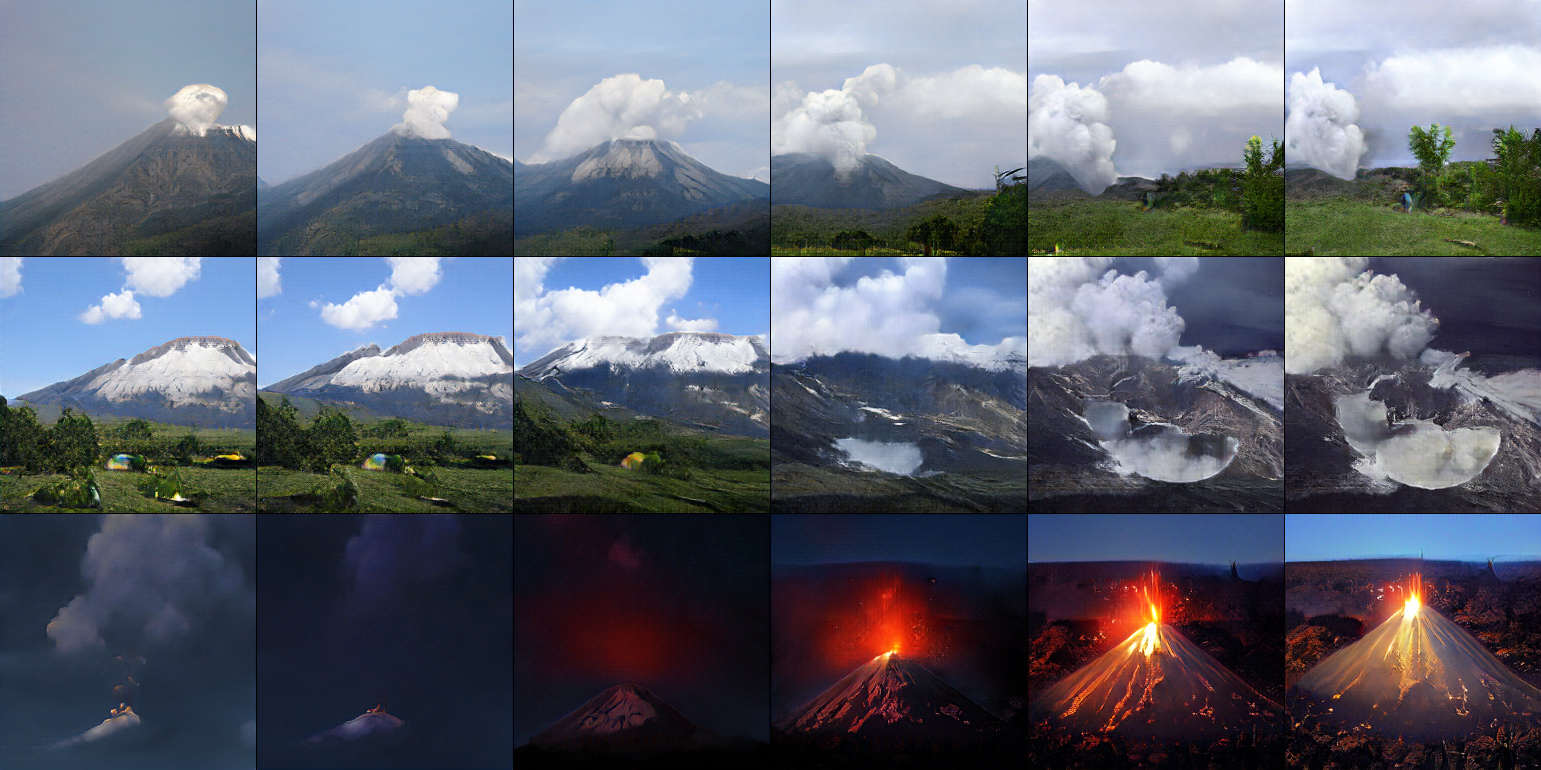}}
\caption{Samples obtained from \tranSSS{} (10\% labels, $256\times256$) when interpolating in the latent space (left to right).\label{fig:s2-20-256-1}}
\end{figure}

\begin{figure}[h]
\centering
\makebox[\textwidth][c]{\includegraphics[scale=1.4]{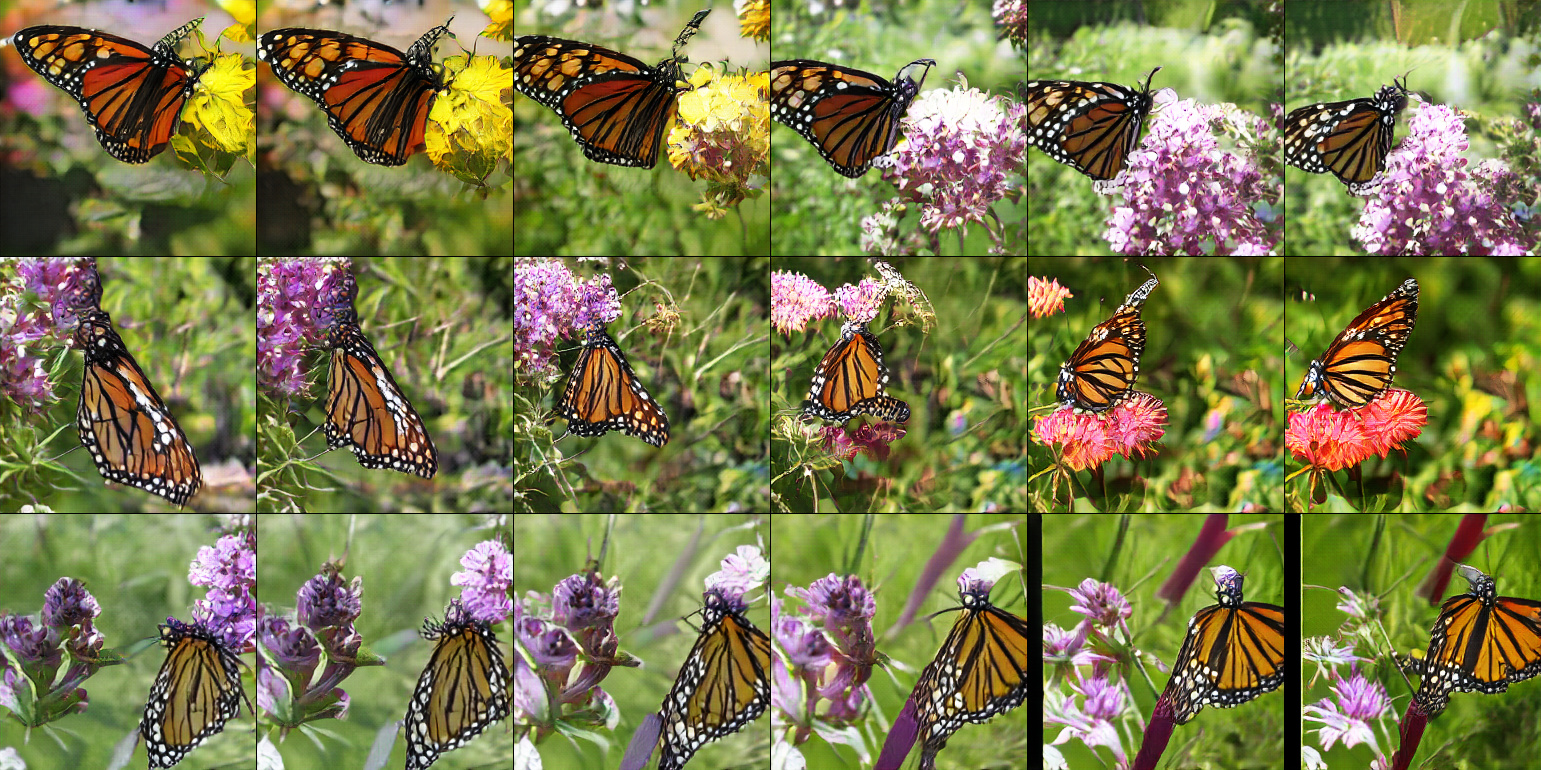}}
\caption{Samples obtained from \tranSSS{} (10\% labels, $256\times256$) when interpolating in the latent space (left to right).\label{fig:s2-20-256-2}}
\end{figure}

\begin{figure*}[h]
    \centering
    \begin{minipage}{0.45\textwidth}
        \centering
        \includegraphics[scale=0.155]{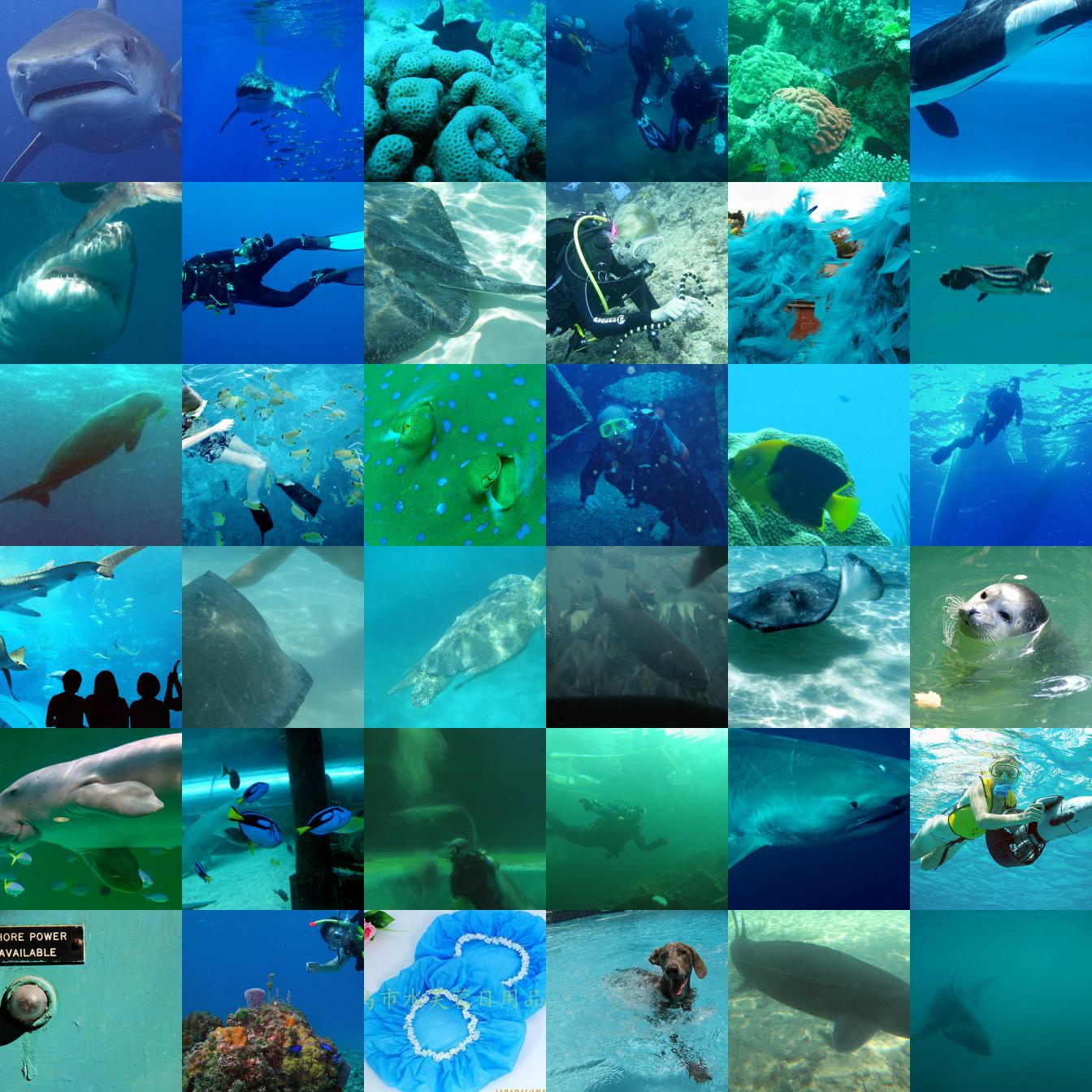}
        {Real images.}
    \end{minipage}
    \begin{minipage}{0.45\textwidth}
        \centering
        \includegraphics[scale=0.15,trim=0 36px 0 36px,clip]{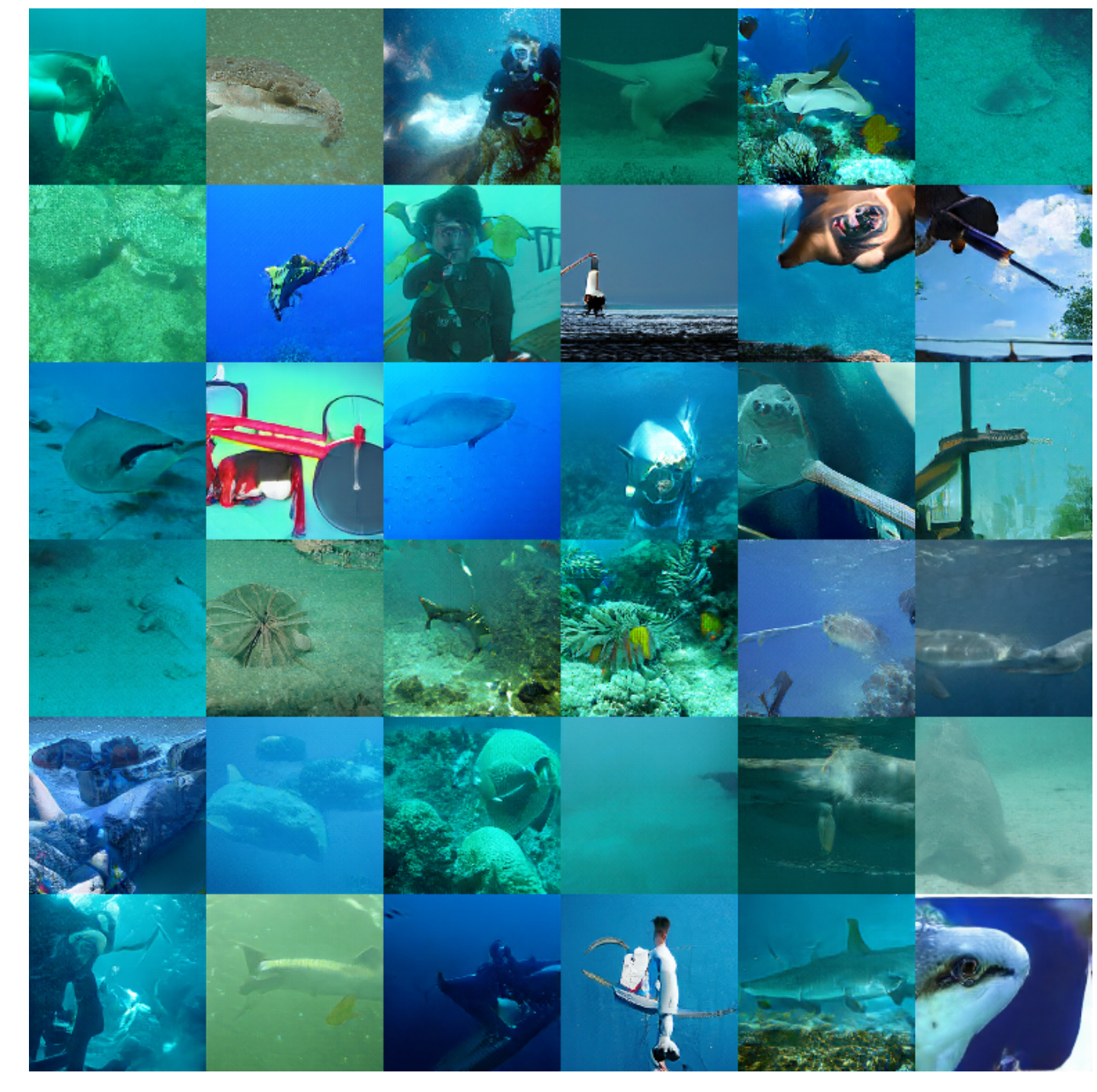}
        {Generated images.}
    \end{minipage}
    \caption{\label{fig:cluster15samples}Real and generated images ($128\times128$) for one of the 50 clusters produced by \tranC{}. Both real and generated images show mostly underwater scenes.}
\end{figure*}

\begin{figure*}
    \centering
    \begin{minipage}{0.45\textwidth}
        \centering
        \includegraphics[scale=0.155]{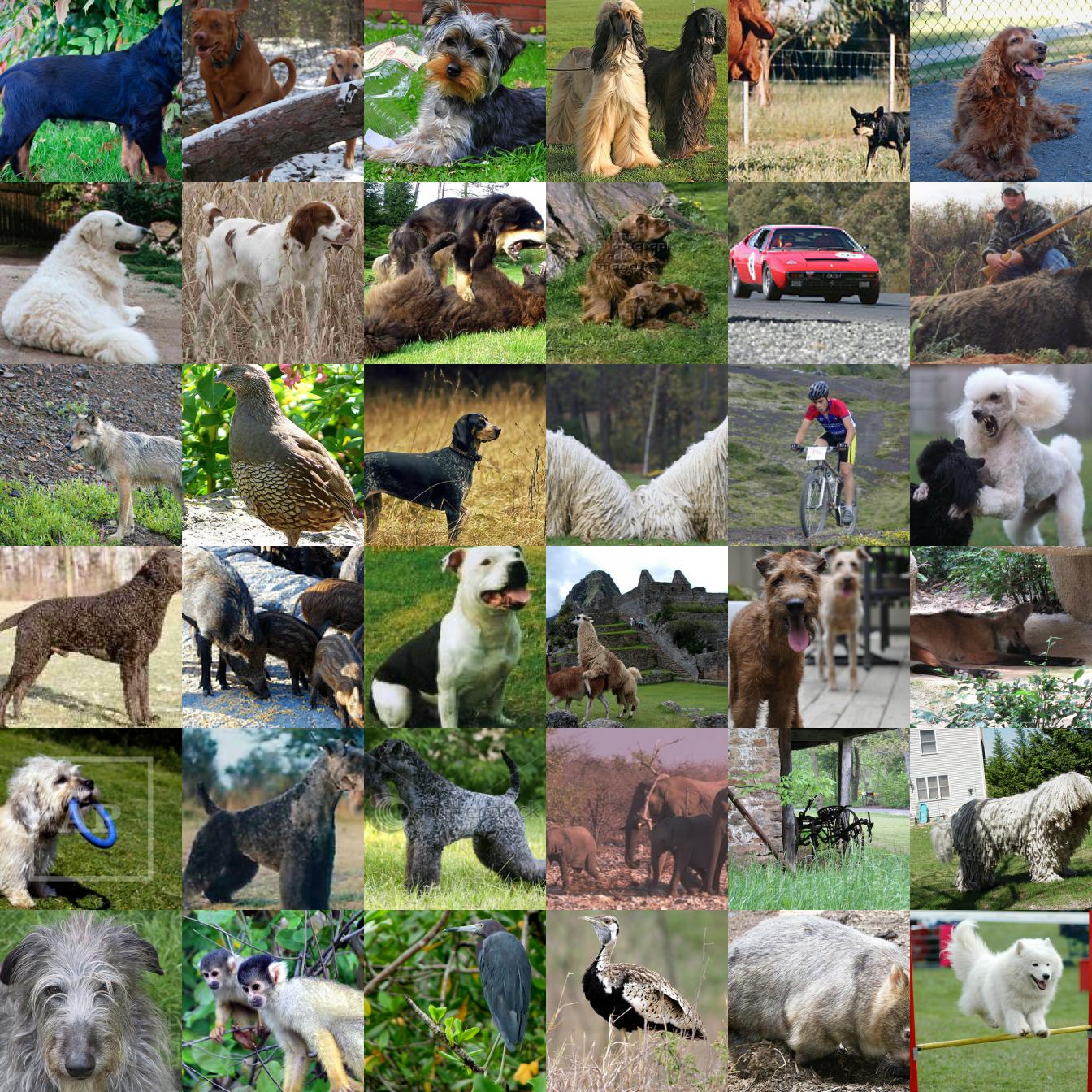}
        {Real images.}
    \end{minipage}
    \begin{minipage}{0.45\textwidth}
        \centering
        \includegraphics[scale=0.15,trim=0 36px 0 36px,clip]{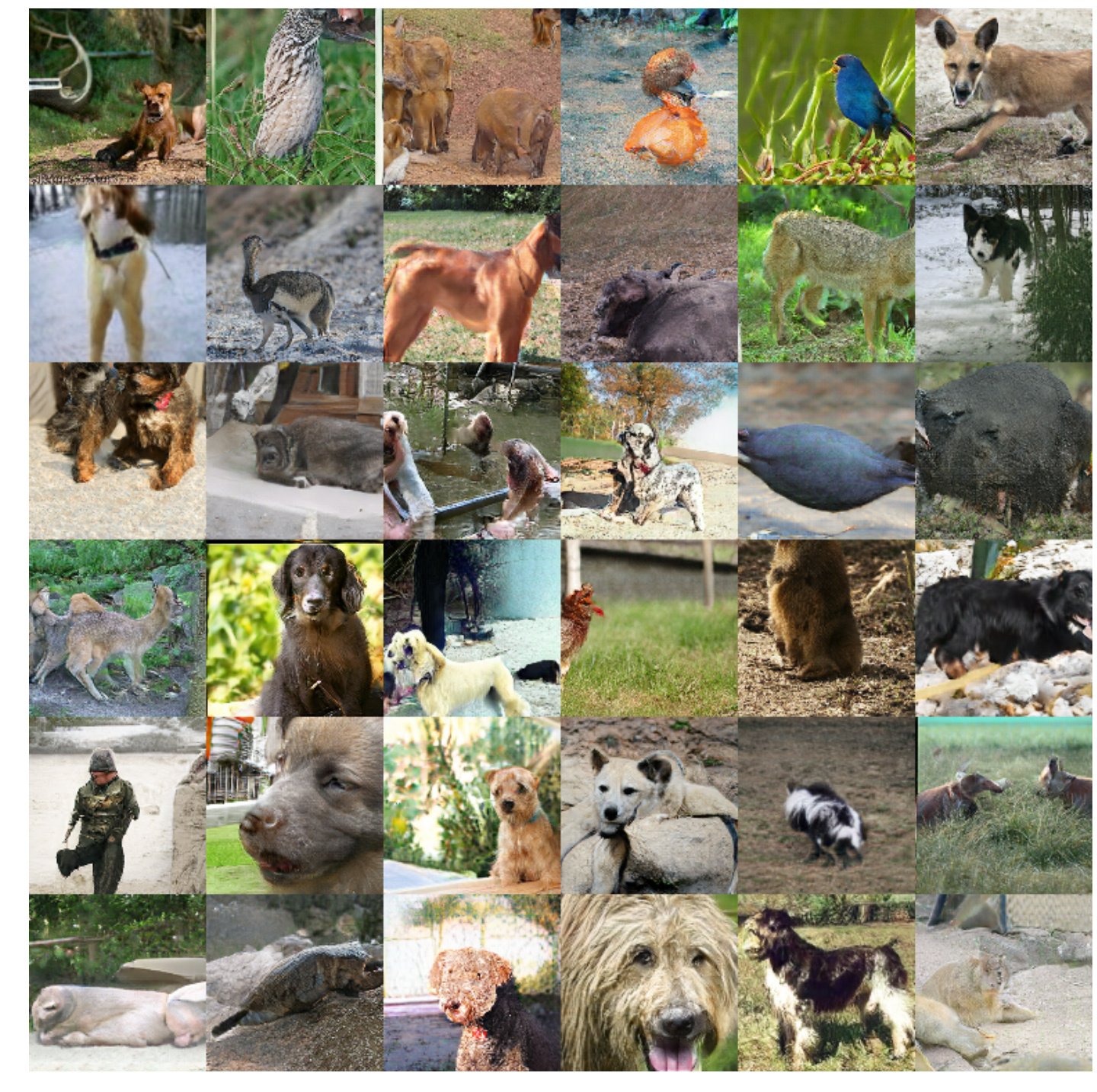}
        {Generated images.}
    \end{minipage}
    \caption{\label{fig:cluster16samples}Real and generated images ($128\times128$) for one of the 50 clusters produced by \tranC{}. Both real and generated images show mostly outdoor scenes featuring different animals.}
\end{figure*}

\begin{figure*}
    \centering
    \begin{minipage}{0.45\textwidth}
        \centering
        \includegraphics[scale=0.155]{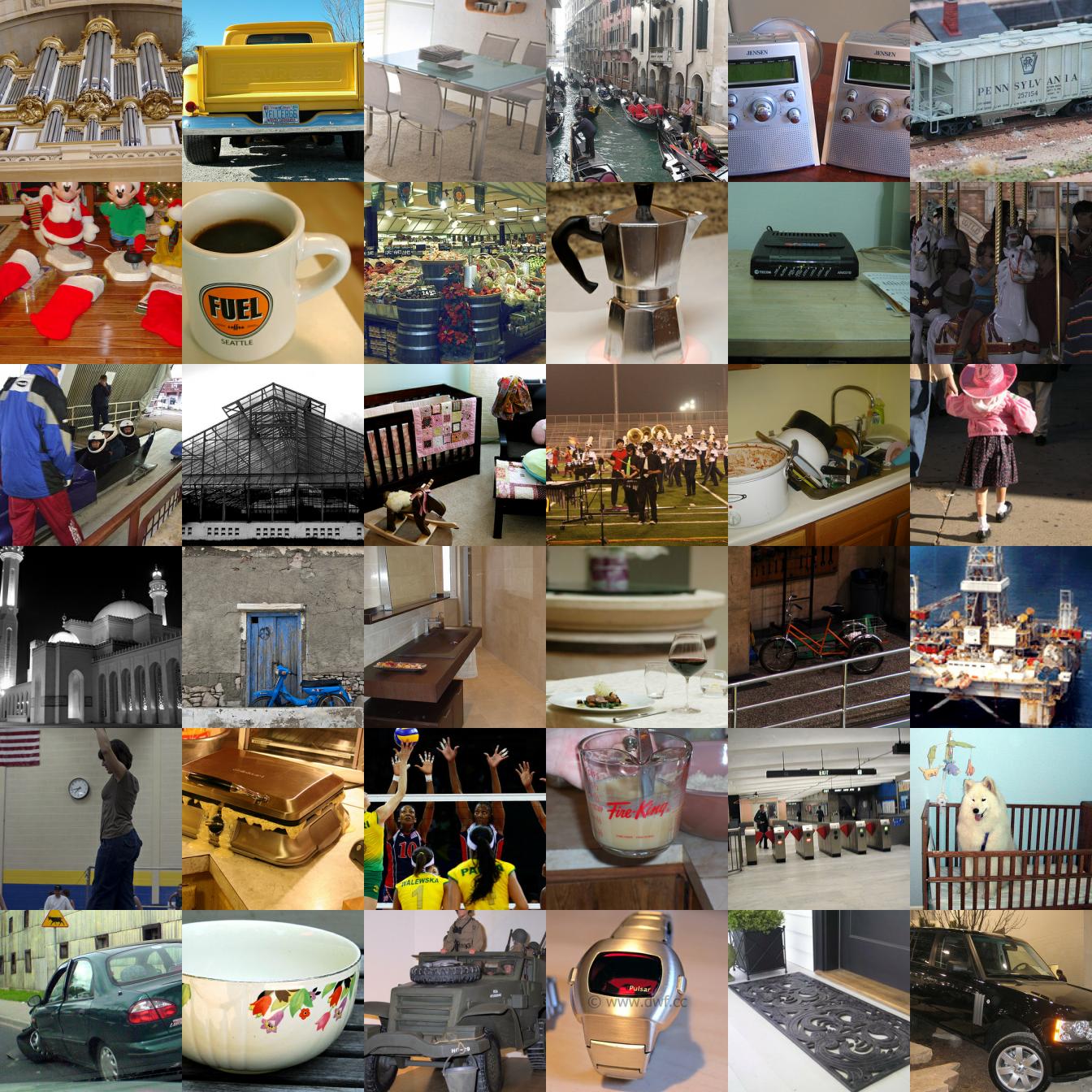}
        {Real images.}
    \end{minipage}
    \begin{minipage}{0.45\textwidth}
        \centering
        \includegraphics[scale=0.15,trim=0 36px 0 36px,clip]{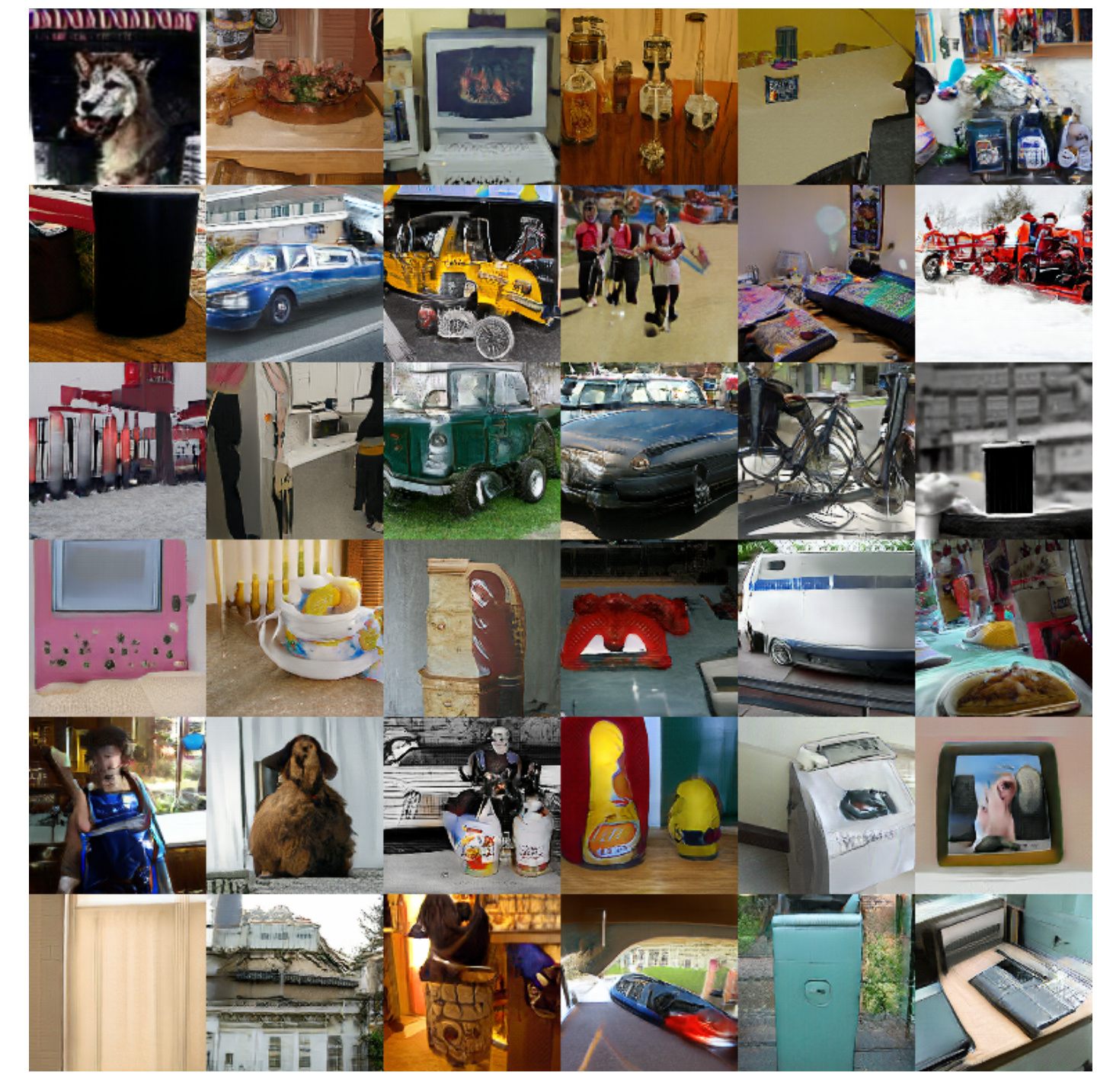}
        {Generated images.}
    \end{minipage}
    \caption{\label{fig:cluster42samples}Real and generated images ($128\times128$) for one of the 50 clusters produced by \tranC{}. In contrast to the examples shown in Figures~\ref{fig:cluster15samples} and \ref{fig:cluster16samples} the clusters show diverse indoor and outdoor scenes.}
\end{figure*}

\begin{figure*}
\centering
\makebox[\textwidth][c]{\includegraphics[scale=0.5]{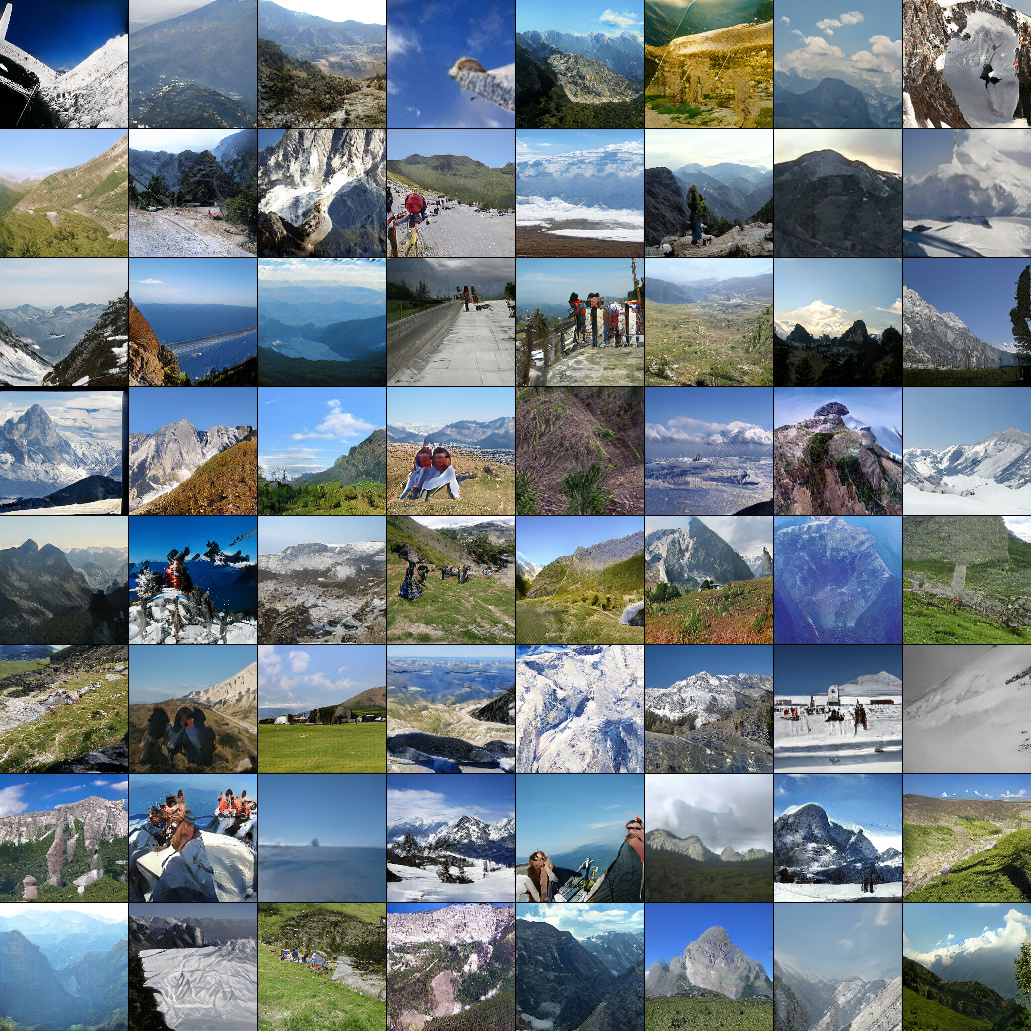}}
\caption{Samples generated by \tranSSS{} (20\% labels, $128\times128$) for a single class. The model captures the great diversity within the class. Human faces and more dynamic scenes present challenges.\label{fig:s2-20-1}}
\end{figure*}

\begin{figure*}
\centering
\makebox[\textwidth][c]{\includegraphics[scale=0.5]{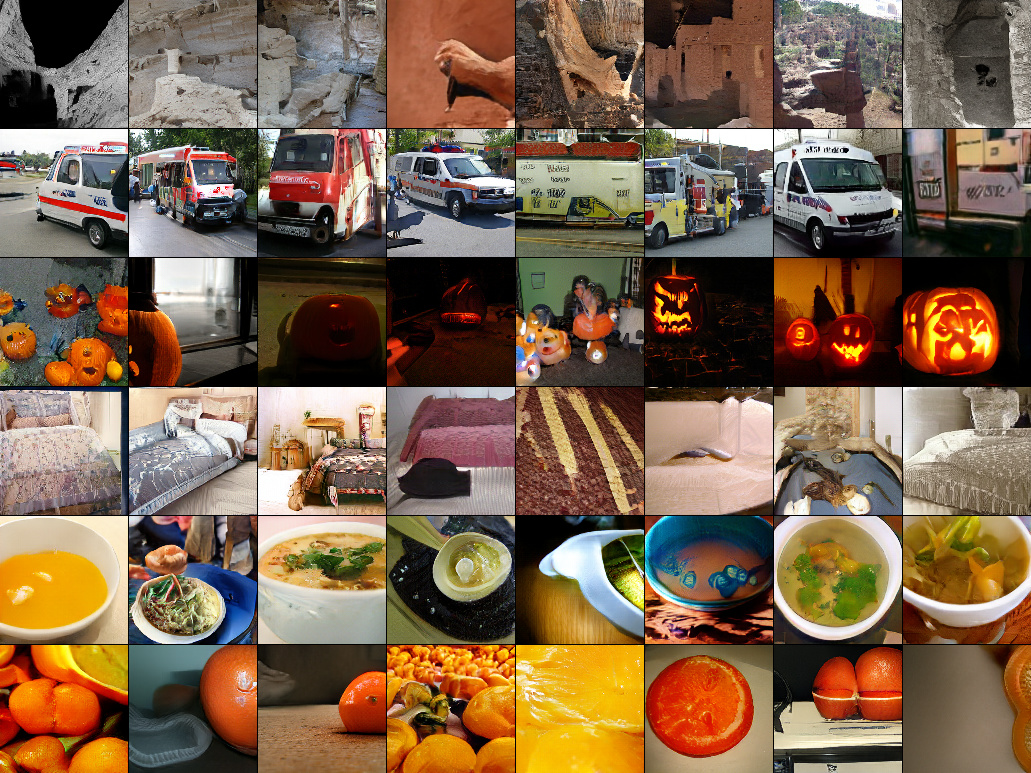}}
\caption{Generated samples by \tranSSS{} (20\% labels, $128\times128$) for different classes. The model correctly learns the different classes and we do not observe class leakage.\label{fig:s2-20-2}}
\end{figure*}

\begin{figure*}
\centering
\includegraphics[scale=1.6]{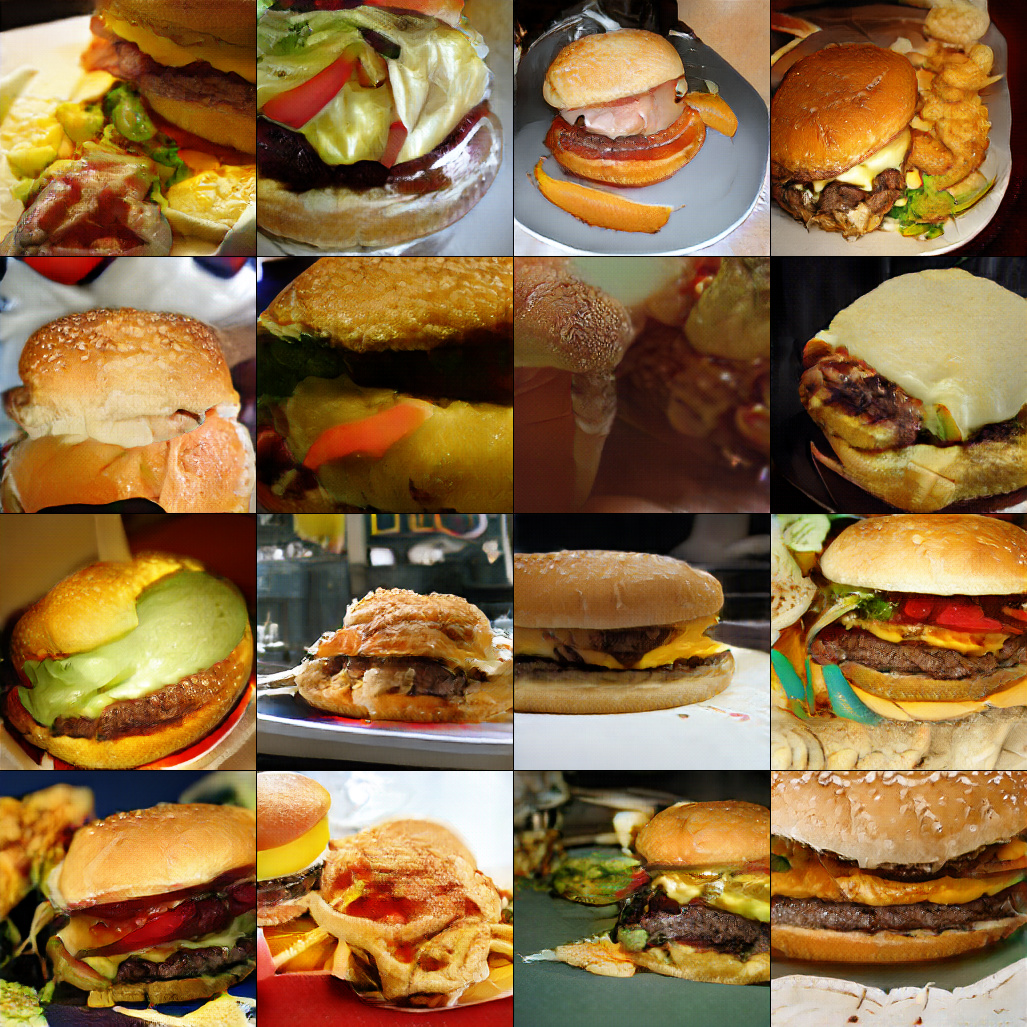}
\caption{Generated samples by \tranSSS{} (10\% labels, $256\times256$) for a single class. The model captures the diversity within the class.\label{fig:s2-20-256-3}}
\end{figure*}

\begin{figure*}
\centering
\includegraphics[scale=1.6]{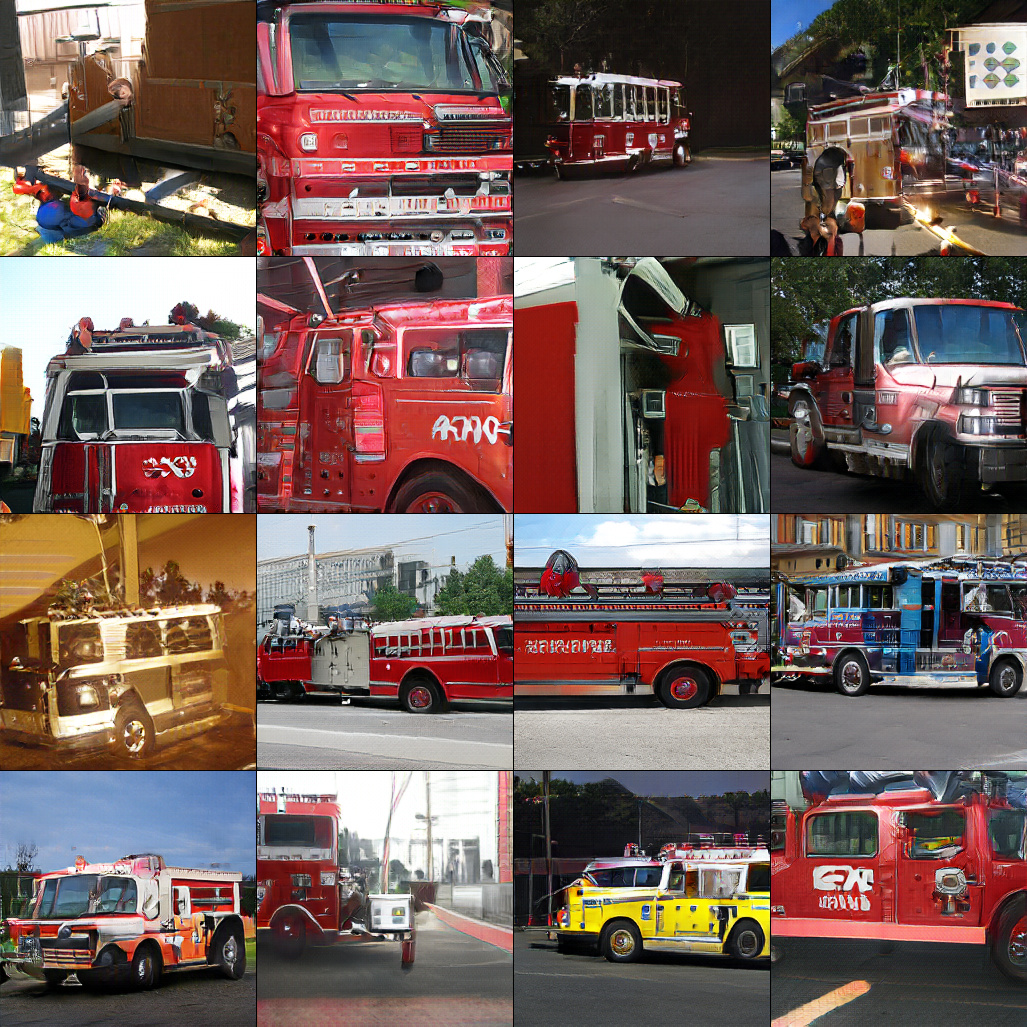}
\caption{Generated samples by \tranSSS{} (10\% labels, $256\times256$) for a single class. The model captures the diversity within the class. \label{fig:s2-20-256-4}}
\end{figure*}

\twocolumn

\section{Architectural details}\label{sec:arch_details}
The ResNet architecture implemented following~\citet{brock2018large} is described in Tables~\ref{tab:resnet_g_blocks} and~\ref{tab:resnet_d_blocks}.
We use the abbreviations RS for resample, BN for batch normalization, and cBN for conditional BN \cite{dumoulin2017learned, de2017modulating}. In the resample column, we indicate downscale(D)/upscale(U)/none(-) setting and in the spectral norm column shows whether spectral normalization is applied to all weights in the layer.
In Table~\ref{tab:resnet_d_blocks}, $y$ stands for the labels and $\bm{h}$ is the output from the layer before (i.e., the pre-logit layer).
Tables~\ref{tab:resblock_generator} and~\ref{tab:resblock_discriminator} show ResBlock details.
The addition layer merges the shortcut path and the convolution path by adding them.
$h$ and $w$ are the input height and width of the ResBlock,
$c_{i}$ and $c_{o}$ are the input channels and output channels for a ResBlock.
For the last ResBlock in the discriminator without downsampling,
we simply drop the shortcut layer from ResBlock. We list all the trainable variables and their shape in Tables~\ref{tab:generator_details} and~\ref{tab:discriminator_details}.
\begin{table}[h]
\centering
\caption{\label{tab:resnet_g_blocks}ResNet generator architecture. ``ch" represents the channel width multiplier and is set to $96$.\vspace{0.2cm}}
\begin{tabular}{lccc}
\toprule
\textsc{Layer}               & \textsc{RS} & \textsc{SN} & \textsc{Output}               \\ \toprule
$\bm z\sim \mathcal{N}(0,1)$ & -           & -           & $120$                         \\ \midrule
Dense                        & -           & -           & $4\times4\times16\cdot ch$    \\
ResBlock                     & U           & SN          & $8\times8\times16\cdot ch$    \\
ResBlock                     & U           & SN          & $16\times16\times8\cdot ch$   \\
ResBlock                     & U           & SN          & $32\times32\times4\cdot ch$   \\
ResBlock                     & U           & SN          & $64\times64\times2\cdot ch$   \\
Non-local block              & -           & -           & $64\times64\times2\cdot ch$   \\
ResBlock                     & U           & SN          & $128\times128\times1\cdot ch$ \\
BN, ReLU                     & -           & -           & $128\times128\times3$         \\
Conv $[3, 3, 1]$             & -           & -           & $128\times128\times3$         \\
Tanh                         & -           & -           & $128\times128\times3$         \\ \bottomrule
\end{tabular}
\end{table}

\begin{table}[h]
 \centering
\caption{\label{tab:resblock_generator}ResBlock generator with upsample.\vspace{0.2cm}}
\begin{tabular}{llll}
  \toprule
      \textsc{Layer} & \textsc{Kernel}& \textsc{RS} & \textsc{Output} \\\toprule
      Shortcut & $[1,1,1]$ & U & $2h \times 2w \times c_{o}$ \\ \midrule
      cBN, ReLU & - & - & $h \times w \times c_{i}$ \\
      Conv & $[3,3,1]$ & U & $2h \times 2w \times c_{o}$ \\
      cBN, ReLU & - & - & $2h \times 2w \times c_{o}$ \\
      Conv & $[3,3,1]$ & - & $2h \times 2w \times c_{o}$ \\ \midrule
      Addition & - & - & $2h \times 2w \times c_{o}$ \\ \bottomrule
    \end{tabular}
\end{table}

\begin{table}[h]
\centering
\caption{\label{tab:resnet_d_blocks}ResNet discriminator architecture. ``ch" represents the channel width multiplier and is set to $96$. Spectral normalization is applied to all layers. \vspace{0.2cm}}
\begin{tabular}{lcc}
\toprule
\textsc{Layer}           & \textsc{RS} & \textsc{Output} \\ \toprule
Input image &  - & $128\times128\times3$            \\\midrule
ResBlock   &   D &     $64\times64\times1 \cdot ch$       \\
Non-local block   &   - &     $64\times64\times1 \cdot ch$       \\
ResBlock   &   D &     $32\times32\times2\cdot ch$       \\
ResBlock   &   D &     $16\times16\times4\cdot ch$       \\
ResBlock   &   D &     $8\times8\times8\cdot ch$       \\
ResBlock   &   D &     $4\times4\times16\cdot ch$       \\
ResBlock (without shortcut)   &   - &     $4\times4\times16\cdot ch$       \\
ReLU & - & $4\times4\times16\cdot ch$       \\
Global sum pooling & - & $1\times1\times16\cdot ch$       \\ \midrule
Sum(embed($y$)$\cdot \bm{h}$)+(dense$\rightarrow 1$) & - &$1$\\ \bottomrule
\end{tabular}
\end{table}

\begin{table}[h]
 \centering
\caption{\label{tab:resblock_discriminator}ResBlock discriminator with downsample.\vspace{0.2cm}}
\begin{tabular}{llll}
  \toprule
      \textsc{Layer} & \textsc{Kernel}& \textsc{RS} & \textsc{Output} \\\toprule
      Shortcut & $[1,1,1]$ & D & $h/2 \times w/2 \times c_{o}$ \\ \midrule
      ReLU & - & - & $h \times w \times c_{i}$ \\
      Conv & $[3,3,1]$ & - & $h \times w \times c_{o}$ \\
      ReLU & - & - & $h \times w \times c_{o}$ \\
      Conv & $[3,3,1]$ & D & $h/2 \times w/2 \times c_{o}$ \\ \midrule
      Addition & - & - & $h/2 \times w/2 \times c_{o}$ \\ \bottomrule
    \end{tabular}
\end{table}

\begin{table*}[h]
    \centering
    \small
    \texttt{
    \input{tables/generator.tex}
    }
    \caption{Tensor-level description of the generator containing a total of 70,433,988 parameters.}
    \label{tab:generator_details}
\end{table*}

\begin{table*}[h]
    \centering
    \small
    \texttt{
    \input{tables/discriminator.tex}
    }
    \caption{Tensor-level description of the discriminator containing a total of 87,982,370 parameters.}
    \label{tab:discriminator_details}
\end{table*}

\FloatBarrier
\onecolumn
\section{FID and IS training curves}
\begin{figure*}[h!]
\centering
\includegraphics[scale=0.49]{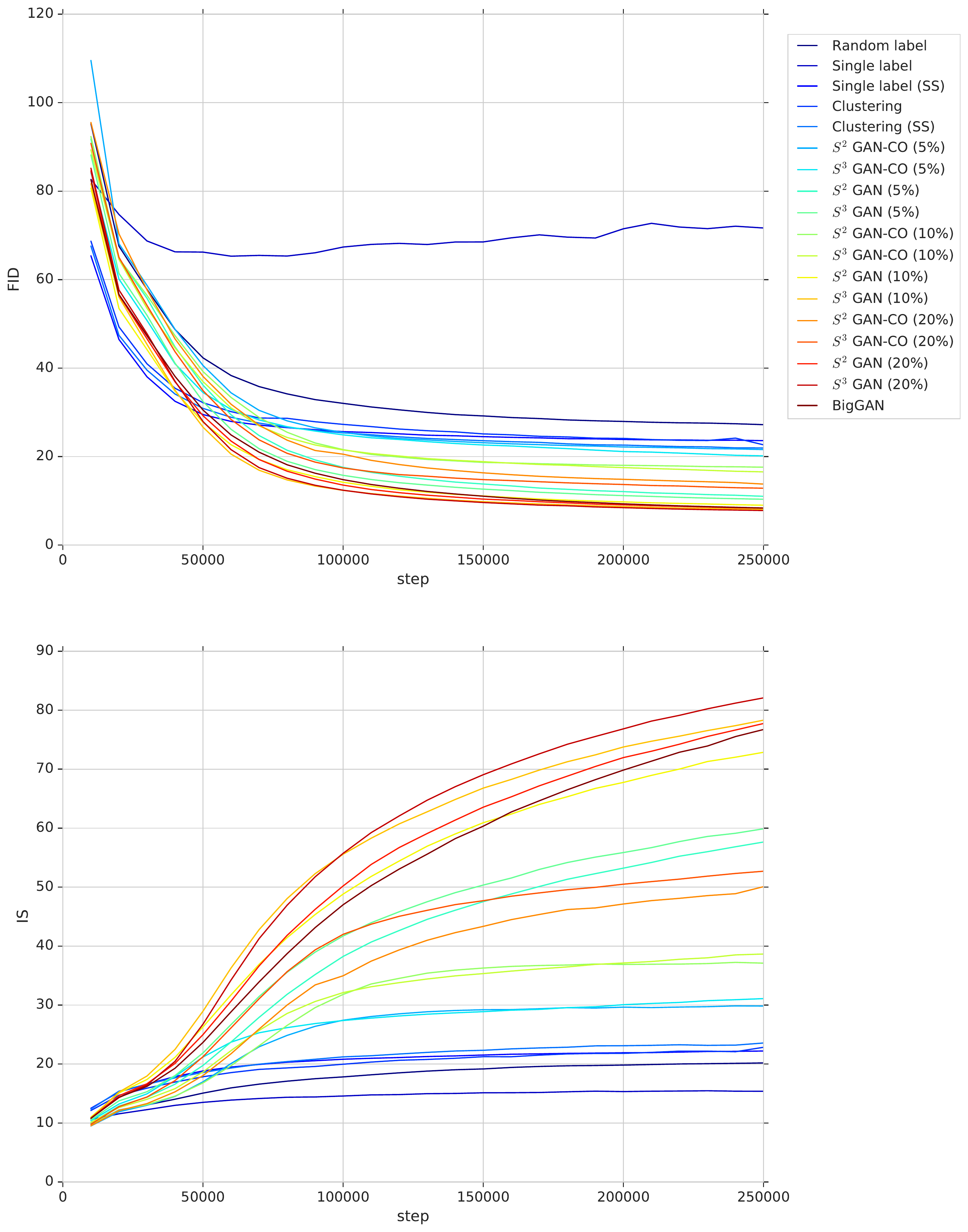}
\caption{Mean FID and IS (3 runs) on ImageNet ($128\times128$) for the models considered in this paper, as a function of the number of generator steps. All models train stably, except \slabels{} (where one run collapsed). \label{fig:convergence}}
\end{figure*}

\FloatBarrier
\onecolumn
\section{FID and IS: Mean and standard deviations} \label{app:meanstd}

\begin{table}[h!]
\centering
\caption{Pre-trained vs co-training approaches, and the effect of self-supervision during GAN training. While co-training approaches outperform fully unsupervised approaches, they are clearly outperformed by the pre-trained approaches. Self-supervision during GAN training helps in all cases.\vspace{0.2cm}}
\label{tab:transfer_vs_direct_mean_std}
\input{tables/transfer_vs_direct_mean_std.tex}
\end{table}

\begin{table}[h!]
\centering
\caption{Training with hard (predicted) labels leads to better models than training with soft (predicted) labels.\vspace{0.2cm}}
\input{tables/soft_vs_hard_mean_std.tex}
\label{tab:soft_vs_hard_mean_std}
\end{table}

\begin{table}[h]
\centering
\caption{Mean FID and IS for the unsupervised approaches.\vspace{0.2cm}}
\label{tab:unsupervised_fid_is_mean_std}
\input{tables/unsupervised_fid_is_mean_std.tex}
\end{table}

\end{document}

%% file: math_commands.tex
\usepackage{amsmath,amsfonts,bm}

\DeclareMathOperator{\FID}{FID} 
\newcommand{\imagenet}{\textsc{imagenet}}

\def\1{\bm{1}}

\DeclareMathAlphabet{\mathsfit}{\encodingdefault}{\sfdefault}{m}{sl}
\SetMathAlphabet{\mathsfit}{bold}{\encodingdefault}{\sfdefault}{bx}{n}

\newcommand{\pdata}{p_{\rm{data}}}

\newcommand{\E}{\mathbb{E}}
\newcommand{\Ls}{\mathcal{L}}

\DeclareMathOperator{\Tr}{Tr}

%% file: tables/unsupervised_fid_is_median.tex
\setlength\tabcolsep{4.5pt}
\begin{tabular}{lrr}
\toprule
{} & \textsc{fid} & \textsc{is} \\\midrule
\rlabels{}        &           26.5 &                 20.2 \\\midrule
\slabels{}        &           25.3 &                 20.4 \\
\slabels{} (SS)   &           23.7 &                 22.2 \\\midrule
\tranC{}          &           23.2 &                 22.7 \\
\tranC{} (SS)     &           22.0 &                 23.5 \\
\bottomrule
\end{tabular}

%% file: tables/semi_self_supervision.tex
\begin{tabular}{lrrr}
\toprule
&\multicolumn{3}{c}{\textsc{Labels}}\\
\textsc{Metric}  &  $5\%$  &    $10\%$ &    $20\%$ \\
\midrule
\textsc{Top-1 error} &  $50.08$ & $36.74$ & $29.21$ \\
\textsc{Top-5 error} &  $26.94$ & $16.04$ & $10.33$ \\
\bottomrule
\end{tabular}

%% file: tables/transfer_vs_direct_median.tex
\setlength\tabcolsep{4.5pt}
\begin{tabular}{lrrrrrr}
\toprule
&\multicolumn{3}{c}{\textsc{FID}} & \multicolumn{3}{c}{\textsc{IS}}\\\cmidrule(lr){2-4}\cmidrule(lr){5-7}
&       $5\%$  &  $10\%$ &   $20\%$ &  $5\%$  &  $10\%$ &   $20\%$ \\
\midrule
\tranSS{}       &           10.8 &   8.9 &   8.4 &  57.6 &  73.4 &  77.4 \\
\cotrainSS	    &           21.8 &  17.7 &  13.9 &  30.0 &  37.2 &  49.2 \\\midrule
\tranSSS{}      &           10.4 &   8.0 &   7.7 &  59.6 &  78.7 &  83.1 \\
\cotrainSSS	    &           20.2 &  16.6 &  12.7 &  31.0 &  38.5 &  53.1 \\
\bottomrule
\end{tabular}

%% file: tables/soft_vs_hard_median.tex
\setlength\tabcolsep{4.5pt}
\begin{tabular}{lrrrrrr}
\toprule
&\multicolumn{3}{c}{\textsc{FID}} & \multicolumn{3}{c}{\textsc{IS}}\\\cmidrule(lr){2-4}\cmidrule(lr){5-7}
&       $5\%$  &  $10\%$ &   $20\%$ &  $5\%$  &  $10\%$ &   $20\%$ \\
\midrule
\tranSS{}       &           10.8 &   8.9 &   8.4 &  57.6 &  73.4 &  77.4 \\
\textsc{+soft}	&           15.4 &  12.9 &  10.4 &  40.3 &  49.8 &  62.1 \\
\bottomrule
\end{tabular}

%% file: tables/generator.tex
\begin{tabular}{lrr}
\toprule
\textsc{Name}                                   &\textsc{Shape}               &\textsc{Size} \\\midrule
generator/embed\_y/kernel:0                               & (1000, 128) &     128,000   \\
generator/fc\_noise/kernel:0                              & (20, 24576) &     491,520   \\
generator/fc\_noise/bias:0                                   & (24576,) &      24,576   \\
generator/B1/bn1/condition/gamma/kernel:0                & (148, 1536) &     227,328   \\
generator/B1/bn1/condition/beta/kernel:0                 & (148, 1536) &     227,328   \\
generator/B1/up\_conv1/kernel:0                    & (3, 3, 1536, 1536) &  21,233,664   \\
generator/B1/up\_conv1/bias:0                                 & (1536,) &       1,536   \\
generator/B1/bn2/condition/gamma/kernel:0                & (148, 1536) &     227,328   \\
generator/B1/bn2/condition/beta/kernel:0                 & (148, 1536) &     227,328   \\
generator/B1/same\_conv2/kernel:0                  & (3, 3, 1536, 1536) &  21,233,664   \\
generator/B1/same\_conv2/bias:0                               & (1536,) &       1,536   \\
generator/B1/up\_conv\_shortcut/kernel:0            & (1, 1, 1536, 1536) &   2,359,296   \\
generator/B1/up\_conv\_shortcut/bias:0                         & (1536,) &       1,536   \\
generator/B2/bn1/condition/gamma/kernel:0                & (148, 1536) &     227,328   \\
generator/B2/bn1/condition/beta/kernel:0                 & (148, 1536) &     227,328   \\
generator/B2/up\_conv1/kernel:0                     & (3, 3, 1536, 768) &  10,616,832   \\
generator/B2/up\_conv1/bias:0                                  & (768,) &         768   \\
generator/B2/bn2/condition/gamma/kernel:0                 & (148, 768) &     113,664   \\
generator/B2/bn2/condition/beta/kernel:0                  & (148, 768) &     113,664   \\
generator/B2/same\_conv2/kernel:0                    & (3, 3, 768, 768) &   5,308,416   \\
generator/B2/same\_conv2/bias:0                                & (768,) &         768   \\
generator/B2/up\_conv\_shortcut/kernel:0             & (1, 1, 1536, 768) &   1,179,648   \\
generator/B2/up\_conv\_shortcut/bias:0                          & (768,) &         768   \\
generator/B3/bn1/condition/gamma/kernel:0                 & (148, 768) &     113,664   \\
generator/B3/bn1/condition/beta/kernel:0                  & (148, 768) &     113,664   \\
generator/B3/up\_conv1/kernel:0                      & (3, 3, 768, 384) &   2,654,208   \\
generator/B3/up\_conv1/bias:0                                  & (384,) &         384   \\
generator/B3/bn2/condition/gamma/kernel:0                 & (148, 384) &      56,832   \\
generator/B3/bn2/condition/beta/kernel:0                  & (148, 384) &      56,832   \\
generator/B3/same\_conv2/kernel:0                    & (3, 3, 384, 384) &   1,327,104   \\
generator/B3/same\_conv2/bias:0                                & (384,) &         384   \\
generator/B3/up\_conv\_shortcut/kernel:0              & (1, 1, 768, 384) &     294,912   \\
generator/B3/up\_conv\_shortcut/bias:0                          & (384,) &         384   \\
generator/B4/bn1/condition/gamma/kernel:0                 & (148, 384) &      56,832   \\
generator/B4/bn1/condition/beta/kernel:0                  & (148, 384) &      56,832   \\
generator/B4/up\_conv1/kernel:0                      & (3, 3, 384, 192) &     663,552   \\
generator/B4/up\_conv1/bias:0                                  & (192,) &         192   \\
generator/B4/bn2/condition/gamma/kernel:0                 & (148, 192) &      28,416   \\
generator/B4/bn2/condition/beta/kernel:0                  & (148, 192) &      28,416   \\
generator/B4/same\_conv2/kernel:0                    & (3, 3, 192, 192) &     331,776   \\
generator/B4/same\_conv2/bias:0                                & (192,) &         192   \\
generator/B4/up\_conv\_shortcut/kernel:0              & (1, 1, 384, 192) &      73,728   \\
generator/B4/up\_conv\_shortcut/bias:0                          & (192,) &         192   \\
generator/non\_local\_block/conv2d\_theta/kernel:0      & (1, 1, 192, 24) &       4,608   \\
generator/non\_local\_block/conv2d\_phi/kernel:0        & (1, 1, 192, 24) &       4,608   \\
generator/non\_local\_block/conv2d\_g/kernel:0          & (1, 1, 192, 96) &      18,432   \\
generator/non\_local\_block/sigma:0                                 & () &           1   \\
generator/non\_local\_block/conv2d\_attn\_g/kernel:0     & (1, 1, 96, 192) &      18,432   \\
generator/B5/bn1/condition/gamma/kernel:0                 & (148, 192) &      28,416   \\
generator/B5/bn1/condition/beta/kernel:0                  & (148, 192) &      28,416   \\
generator/B5/up\_conv1/kernel:0                       & (3, 3, 192, 96) &     165,888   \\
generator/B5/up\_conv1/bias:0                                   & (96,) &          96   \\
generator/B5/bn2/condition/gamma/kernel:0                  & (148, 96) &      14,208   \\
generator/B5/bn2/condition/beta/kernel:0                   & (148, 96) &      14,208   \\
generator/B5/same\_conv2/kernel:0                      & (3, 3, 96, 96) &      82,944   \\
generator/B5/same\_conv2/bias:0                                 & (96,) &          96   \\
generator/B5/up\_conv\_shortcut/kernel:0               & (1, 1, 192, 96) &      18,432   \\
generator/B5/up\_conv\_shortcut/bias:0                           & (96,) &          96   \\
generator/final\_norm/gamma:0                                   & (96,) &          96   \\
generator/final\_norm/beta:0                                    & (96,) &          96   \\
generator/final\_conv/kernel:0                          & (3, 3, 96, 3) &       2,592   \\
generator/final\_conv/bias:0                                     & (3,) &           3   \\
\bottomrule
\end{tabular}

%% file: tables/discriminator.tex
\begin{tabular}{lrr}
\toprule
\textsc{Name}                                   &\textsc{Shape}               &\textsc{Size} \\\midrule
discriminator/B1/same\_conv1/kernel:0			& (3, 3, 3, 96)&       2,592   \\
discriminator/B1/same\_conv1/bias:0                      & (96,)&          96   \\
discriminator/B1/down\_conv2/kernel:0			& (3, 3, 96, 96)&      82,944   \\
discriminator/B1/down\_conv2/bias:0                      & (96,)&          96   \\
discriminator/B1/down\_conv\_shortcut/kernel:0		& (1, 1, 3, 96)&         288   \\
discriminator/B1/down\_conv\_shortcut/bias:0             & (96,)&          96   \\
discriminator/non\_local\_block/conv2d\_theta/kernel:0	& (1, 1, 96, 12)&       1,152   \\
discriminator/non\_local\_block/conv2d\_phi/kernel:0	& (1, 1, 96, 12)&       1,152   \\
discriminator/non\_local\_block/conv2d\_g/kernel:0	& (1, 1, 96, 48)&       4,608   \\
discriminator/non\_local\_block/sigma:0                  & ()&           1   \\
discriminator/non\_local\_block/conv2d\_attn\_g/kernel:0	& (1, 1, 48, 96)&       4,608   \\
discriminator/B2/same\_conv1/kernel:0			& (3, 3, 96, 192)&     165,888   \\
discriminator/B2/same\_conv1/bias:0                      & (192,)&         192   \\
discriminator/B2/down\_conv2/kernel:0			& (3, 3, 192, 192)&     331,776   \\
discriminator/B2/down\_conv2/bias:0                      & (192,)&         192   \\
discriminator/B2/down\_conv\_shortcut/kernel:0		& (1, 1, 96, 192)&      18,432   \\
discriminator/B2/down\_conv\_shortcut/bias:0             & (192,)&         192   \\
discriminator/B3/same\_conv1/kernel:0			& (3, 3, 192, 384)&     663,552   \\
discriminator/B3/same\_conv1/bias:0                      & (384,)&         384   \\
discriminator/B3/down\_conv2/kernel:0			& (3, 3, 384, 384)&   1,327,104   \\
discriminator/B3/down\_conv2/bias:0                      & (384,)&         384   \\
discriminator/B3/down\_conv\_shortcut/kernel:0		& (1, 1, 192, 384)&      73,728   \\
discriminator/B3/down\_conv\_shortcut/bias:0             & (384,)&         384   \\
discriminator/B4/same\_conv1/kernel:0			& (3, 3, 384, 768)&   2,654,208   \\
discriminator/B4/same\_conv1/bias:0                      & (768,)&         768   \\
discriminator/B4/down\_conv2/kernel:0			& (3, 3, 768, 768)&   5,308,416   \\
discriminator/B4/down\_conv2/bias:0                      & (768,)&         768   \\
discriminator/B4/down\_conv\_shortcut/kernel:0		& (1, 1, 384, 768)&     294,912   \\
discriminator/B4/down\_conv\_shortcut/bias:0             & (768,)&         768   \\
discriminator/B5/same\_conv1/kernel:0			& (3, 3, 768, 1536)&  10,616,832   \\
discriminator/B5/same\_conv1/bias:0                      & (1536,)&       1,536   \\
discriminator/B5/down\_conv2/kernel:0			& (3, 3, 1536, 1536)&  21,233,664   \\
discriminator/B5/down\_conv2/bias:0                      & (1536,)&       1,536   \\
discriminator/B5/down\_conv\_shortcut/kernel:0		& (1, 1, 768, 1536)&   1,179,648   \\
discriminator/B5/down\_conv\_shortcut/bias:0             & (1536,)&       1,536   \\
discriminator/B6/same\_conv1/kernel:0			& (3, 3, 1536, 1536)&  21,233,664   \\
discriminator/B6/same\_conv1/bias:0                      & (1536,)&       1,536   \\
discriminator/B6/same\_conv2/kernel:0			& (3, 3, 1536, 1536)&  21,233,664   \\
discriminator/B6/same\_conv2/bias:0                      & (1536,)&       1,536   \\
discriminator/final\_fc/kernel:0                         & (1536, 1)&       1,536   \\
discriminator/final\_fc/bias:0                           & (1,)&           1   \\
discriminator\_projection/kernel:0			& (1000, 1536)&   1,536,000   \\
\bottomrule
\end{tabular}

%% file: tables/transfer_vs_direct_mean_std.tex
\setlength\tabcolsep{4.5pt}
\begin{tabular}{lrrrrrr}
\toprule
&\multicolumn{3}{c}{\textsc{FID}} & \multicolumn{3}{c}{\textsc{IS}}\\\cmidrule(lr){2-4}\cmidrule(lr){5-7}
&       $5\%$  &  $10\%$ &   $20\%$ &  $5\%$  &  $10\%$ &   $20\%$ \\\midrule
\tranSS{}           &  11.0$\pm$0.31 &   9.0$\pm$0.30 &   8.4$\pm$0.02 &  57.6$\pm$0.86 &  72.9$\pm$1.41 &  77.7$\pm$1.24 \\
\cotrainSS{}  &  21.6$\pm$0.64 &  17.6$\pm$0.27 &  13.8$\pm$0.48 &  29.8$\pm$0.21 &  37.1$\pm$0.54 &  50.1$\pm$1.45 \\
\tranSSS{}           &  10.3$\pm$0.16 &   8.1$\pm$0.14 &   7.8$\pm$0.20 &  59.9$\pm$0.74 &  78.3$\pm$1.08 &  82.1$\pm$1.89 \\
\cotrainSSS{}  &  20.2$\pm$0.14 &  16.5$\pm$0.12 &  12.8$\pm$0.51 &  31.1$\pm$0.18 &  38.7$\pm$0.36 &  52.7$\pm$1.08 \\
\bottomrule
\end{tabular}

%% file: tables/soft_vs_hard_mean_std.tex
\setlength\tabcolsep{4.5pt}
\begin{tabular}{lrrrrrr}
\toprule
&\multicolumn{3}{c}{\textsc{FID}} & \multicolumn{3}{c}{\textsc{IS}}\\\cmidrule(lr){2-4}\cmidrule(lr){5-7}
&       $5\%$  &  $10\%$ &   $20\%$ &  $5\%$  &  $10\%$ &   $20\%$ \\
\midrule
\tranSS{}       &  11.0$\pm$0.31 &   9.0$\pm$0.30 &   8.4$\pm$0.02 &  57.6$\pm$0.86 &  72.9$\pm$1.41 &  77.7$\pm$1.24 \\
\tranSS{} \textsc{Soft}  &  15.6$\pm$0.58 &  13.3$\pm$1.71 &  11.3$\pm$1.42 &  40.1$\pm$0.97 &  49.3$\pm$4.67 &  58.5$\pm$5.84 \\
\bottomrule
\end{tabular}

%% file: tables/unsupervised_fid_is_mean_std.tex
\setlength\tabcolsep{4.5pt}
\begin{tabular}{lrr}
\toprule
{} & \textsc{fid} & \textsc{IS} \\\midrule
\tranC     &   22.7$\pm$0.80 &  22.8$\pm$0.42 \\
\tranC (SS) &   21.9$\pm$0.08 &  23.6$\pm$0.19 \\
\rlabels          &   27.2$\pm$1.46 &  20.2$\pm$0.33 \\
\slabels          &  71.7$\pm$66.32 &  15.4$\pm$7.57 \\
\slabels (SS)     &   23.6$\pm$0.14 &  22.2$\pm$0.10 \\
\bottomrule
\end{tabular}